\documentclass[10pt]{article} % For LaTeX2e
\usepackage[accepted]{rlc}
\usepackage{amsmath,amsfonts,bm}

\def\eqref#1{equation~\ref{#1}}
\def\1{\bm{1}}

\def\vone{{\bm{1}}}

\DeclareMathAlphabet{\mathsfit}{\encodingdefault}{\sfdefault}{m}{sl}
\SetMathAlphabet{\mathsfit}{bold}{\encodingdefault}{\sfdefault}{bx}{n}

\def\gM{{\mathcal{M}}}
\def\gN{{\mathcal{N}}}
\def\gO{{\mathcal{O}}}
\def\gP{{\mathcal{P}}}

\newcommand{\squareb}[1]{\left[ #1 \right]}

\newcommand{\EEX}[2]{\mathbb{E}_{#1}\squareb{ #2}}

\newcommand{\ltwonorm}[1]{\left\lVert #1 \right\rVert_2}

\newcommand{\ddt}[1]{\frac{\mathrm{d}}{\mathrm{d} #1}}

\newcommand{\vecR}[2]{\mathbb{R}^{#1 \times #2}}

\usepackage{amssymb}            % Defines common symbols like \mathbb R
\usepackage{mathtools}          % Extends amsmath, providing common math tools
\usepackage{mathrsfs}           % Enables \mathscr, which can work in cases that \mathcal does not
\mathtoolsset{showonlyrefs}     % Only number equations that are referenced (optional)
\usepackage{graphicx}           % For including images
\usepackage{subcaption}         % Allows for the use of subfigures and subcaptions
\usepackage[space]{grffile}     % For spaces in image names
\usepackage{url}                % For displaying urls
\usepackage{todonotes}
\usepackage{bm}
\usepackage{bbm}
\usepackage{multirow}
\usepackage{wrapfig}
\usepackage{svg}
\usepackage{mathtools}

\usepackage{amsthm}

\usepackage[many]{tcolorbox}

\usepackage[capitalize,nameinlink,noabbrev]{cleveref} 
\crefformat{equation}{(#2#1#3)}

\usepackage{thm-restate}

\usepackage{uoftcolors}

\newtheorem{assumption}{Assumption}

\newtheorem{lemma}{Lemma}

\newtheorem{insight}{Insight}
\newtheorem{definition}{Definition}

\AtBeginDocument{}%
\AtBeginDocument{}%

\title{When does Self-Prediction help? Understanding 
Auxiliary Tasks in Reinforcement Learning}

\author{%
Claas Voelcker\\
University of Toronto\\
Vector Institute, Toronto\\
cvoelcker@cs.toronto.edu \\\And
Tyler Kastner\\
University of Toronto\\
Vector Institute, Toronto\\
tkastner@cs.toronto.edu \\\And
Igor Gilitschenski\\
University of Toronto\\
Vector Institute, Toronto\\
gilitschenski@cs.toronto.edu\\\AND
\And
Amir-massoud Farahmand\\
University of Toronto\\
farahmand@cs.toronto.edu}
\begin{document}

\maketitle

\begin{abstract}
We investigate the impact of auxiliary learning tasks such as observation reconstruction and latent self-prediction on the representation learning problem in reinforcement learning.
We also study how they interact with distractions and observation functions in the MDP.
We provide a theoretical analysis of the learning dynamics of observation reconstruction, latent self-prediction, and TD learning in the presence of distractions and observation functions under linear model assumptions.
With this formalization, we are able to explain why latent-self prediction is a helpful \emph{auxiliary task}, while observation reconstruction can provide more useful features when used in isolation.
Our empirical analysis shows that the insights obtained from our learning dynamics framework predicts the behavior of these loss functions beyond the linear model assumption in non-linear neural networks.
This reinforces the usefulness of the linear model framework not only for theoretical analysis, but also practical benefit for applied problems.
\end{abstract}

\section{Introduction}

Since the emergence of deep learning, techniques for deep supervised learning have been successfully incorporated into reinforcement learning (RL) agents \citep{dqn,ddpg}.
However, the RL setting contains additional complications such as non-stationary optimization target and the reliance on bootstrapping.
These hurdles generally add instability to the RL training process, and recent work has identified the \emph{failure to learn good features} as a central problem in deep RL \citep{lyle2022understanding,nikishin2022primacy,kumar2021implicit}.

To mitigate this failure, one common approach is to add auxiliary tasks to the learning objective \citep{jaderberg2017reinforcement}. 
Popular examples include predicting next state observations \citep{jaderberg2016reinforcement} and predicting functions of the next state \citep{schwarzer2021dataefficient,ni2024bridging}.
To understand the performance of these approaches, recent literature \citep{tang2022understanding,lelan2023bootstrapped} considers the \emph{learning dynamics} of auxiliary task learning in simple linear surrogate models \citep{saxe2014exact}.
One hypothesis in the literature is that observation reconstruction should provide better features than latent self-prediction \citep{behzadian2019fast,tang2022understanding}. 
However, this causes a theory-practice gap as empirical work has found that latent self-prediction outperforms observation reconstruction across many benchmarks \citep{schwarzer2021dataefficient,ni2024bridging}.

To address this gap, we pose two questions: \emph{(a) How do auxiliary losses behave when combined with a TD loss?} \citet{tang2022understanding,lelan2022generalization,tang2023towards} have studied the learning dynamics of auxiliary tasks alone, evaluating their performance without addressing the interaction between the auxiliary task and the main goal, to learn a (correct) value function. \emph{(b) How can we describe the behavior of auxiliary losses in the presence of distractions (states and transition dynamics irrelevant for the reward) and observation functions (different ways to measure the underlying state)?} MDP structures like distractions have been hypothesized to lead to differing performance between different auxiliary tasks \citep{ni2024bridging}, but to our knowledge no theoretical study has been established.

In \autoref{sec:formalism}, we present a formalization of distractions and observation functions. We use the framework of \emph{factored MDPs} \citep{boutilier2000stochastic} with Kronecker products \citep{mahadevan2009learning} to represent a common class of distractions. To model observation functions, we use \emph{linear reparametrization} as a tractable way to go beyond one-hot representations. 

\begin{figure}
    \centering
    \includegraphics[width=\textwidth]{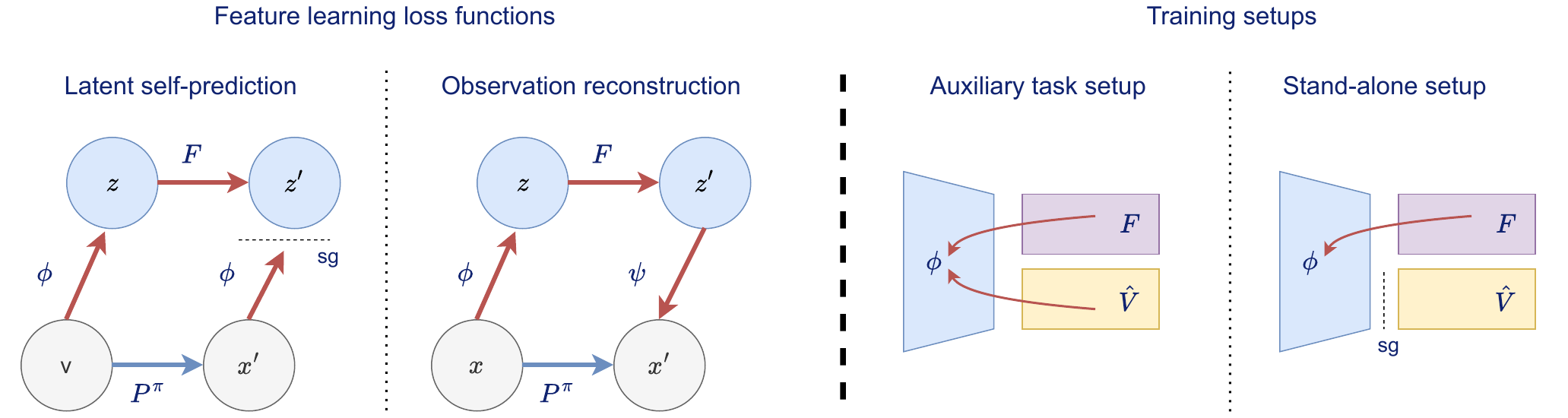}
    \caption{Diagram of the considered loss functions and the different use cases. In latent self-prediction, the aim is to predict next state \emph{features} given by a embedding function $\phi(x')$ using the states features $\phi(x)$ and a latent prediction model $F$. In observation reconstruction, the aim is to match next state \emph{ground truth observations} $x'$ via the use of a decoder function $\psi(F(\phi(x)))$. In the \emph{auxiliary task setup}, both the gradients from the feature learning loss and value function learning are propagated to the encoder, while in the \emph{stand-alone scenario}, only the gradients from the feature learning loss are used to update $\Phi$.}
    \label{fig:losses}
\end{figure}

In \autoref{sec:stand_alone_tasks} and \autoref{sec:auxilliary_tasks} we analyze the features learned with observation prediction and latent self-prediction alone and in combination with TD learning. 
We also show how these stationary features change with the introduction of distractions and observation functions.
From this analysis we find that latent self-prediction is a strong \emph{auxiliary task}, while observation prediction is a strong feature learning method when used \emph{alone}.
The differences are highlighted in \autoref{fig:losses}.
This bridges one of the biggest gaps between previous analysis of learning dynamics and empirical results.

In \autoref{sec:empirical} we test the predictions derived from our theoretical framework by evaluating feature learning losses in the MinAtar suite \citep{young19minatar}.
We design ablations that mirror both our formalization and previous approaches to test distraction robustness in empirical environments. 
The theory partially predicts the performance differences in the test suite, validating that the insights we obtain from the simple linear surrogate models used for analysis are useful for practitioners.
However, we also find surprising deviations from our predictions on some environments, suggesting that there is need for additional research to fully bridge the theory-practice gap.

\section{Background}
We briefly introduce the standard formalism of reinforcement learning and linear value function approximations to clarify the notation used.
Following this, we briefly introduce the training dynamics framework of \citet{tang2022understanding,lelan2023bootstrapped}. 
We frame all losses analyzed in this work with the same models to highlight similarities and differences.

\textbf{Reinforcement Learning.} We consider a discounted Markov decision process (MDP)~\citep{Puterman1994MarkovDP,suttonbook} $(\mathcal{X}, \mathcal{A}, \gP, r, \gamma)$, with state space $\mathcal{X}$, action space $\mathcal{A}$, transition kernel $\gP:\mathcal{X} \times \mathcal{A} \times \mathcal{X} \to \mathbb{R}$,
reward function $r: \mathcal{X} \times \mathcal{A} \to \mathbb{R}$, and discount factor $\gamma \in [0,1)$. Given a policy $\pi:\mathcal{X}\times \mathcal{A} \to \mathbb{R}$, the value function is defined as the expected return conditioned on a state $x$ $$V^\pi(x) = \EEX{\pi}{\sum_{t \geq 0} \gamma^t r_t|x_0 = x},$$ where $r_t = r(x_t, a_t)$ is the reward at time $t$. The goal of an agent is to maximize its value at each state, using the (approximate) value function of its policy $V^\pi$.

In finite state spaces ($\lvert\mathcal{X}\rvert = n$ for some integer $n$), the policy induced transition kernel $\gP^\pi(x'|x) = \int \mathcal{P}(x'|x,a) \mathrm{d}\pi(a|x)$ can be represented as a stochastic matrix $P^\pi$, and the reward function as a vector $r^\pi$.
This permits the compact notation $V^\pi =(I - \gamma P^\pi)^{-1} r^\pi$.
We review several additional properties of stochastic matrices in \autoref{app:linalg}.

\textbf{Value function approximation.} A finite state value function can always be expressed as a table or vector, but it is often infeasible to do so, due to limited storage capacity or resources once the state space becomes sufficiently large.
Therefore, function approximations need to be introduced, which commonly take the form of a feature function $\phi(x)$ and a weight vector $\hat{V}$, with $V^\pi(x) \approx \phi(x)^\top \hat{V}$.
These features can be pre-specified or learned, i.e. using neural networks.
Since we are interested in finite state spaces and linear models, we will assume that $\phi(x): \mathbb{R}^{n} \rightarrow \mathbb{R}^{k}$ with $k < n$  represented by a matrix $\Phi \in \mathbb{R}^{n\times k}$ and $\hat{V} \in \mathbb{R}^k$.

\subsection{Two-layer linear networks as analytical models for training dynamics}
\label{sec:background}

Rigorously analyzing the effect of different loss functions on neural networks is challenging due to non-linearities in the networks, shifting data distributions, and policy updates.
Therefore, we have to resort to studying simplified models to obtain quantitative and qualitative results, and only consider the fixed policy case in our analysis\footnote{We discuss these and other assumptions and their implications in detail in \autoref{app:limitations}.}.
Studying feature learning dynamics using linear networks was popularized by \citet{saxe2014exact} and has proven to be a valuable tool to analyze diverse objectives such as TD learning \citep{tang2023towards,lelan2023bootstrapped}, latent self-prediction \citep{tian2021understanding,tang2022understanding}, and linear autoencoders \citep{pretorius2018learning,bao2020regularized}.

We rewrite the feature learning algorithms using two to three matrices in lieu of more complex functions. Furthermore, we use several assumptions throughout the paper that are listed here for clarity.

\begin{assumption}
\label{assumption1}
Let $\Phi \in \mathbb{R}^{n \times k}$ be an \emph{encoder} mapping to a $k$ dimensional embedding space,
$F \in \mathbb{R}^{k\times k}$ a \emph{latent model} mapping to the next state's latent embedding, $\hat{V} \in \mathbb{R}^{k}$ and $\hat{r} \in \mathbb{R}^{k}$ \emph{value and reward weights}, and $\Psi \in \mathbb{R}^{k \times n}$ a \emph{decoder}.
Let the sampling distribution of state samples $\mathcal{D}$ be uniform and fixed throughout learning.
\end{assumption}

Using this notation, we study four important loss functions for RL: observation reconstruction, where the aim is to fit the next state observation $x^\top \Phi F \Psi \approx x'$, latent reconstruction $x^\top \Phi F \approx x' \Phi$, where the aim is to predict learned features of the next state, and TD learning $x^\top \Phi \hat{V} \approx x^\top(r^\pi + \gamma x'^\top\phi \hat{V})$. To clarify the differences, we show a diagram explaining the losses and training setups in \autoref{fig:losses}.
Following common notation, $[\cdot]_\mathrm{sg}$ signifies a \emph{stop-gradient} operation; no gradient is taken with regard to terms in the parenthesis.

Formally, these are written as
\begin{align*}
    \text{Reconstruction: }&& L_{\text{rec}}(\Phi,F,\Psi) =& \,\EEX{x \sim \mathcal{D}}{\ltwonorm{x^\top \Phi F \Psi - x^\top {P^\pi}}^2},\\
    \text{Latent self-prediction: }&& L_{\text{lat}}(\Phi,F) =& \,\EEX{x \sim \mathcal{D}}{\ltwonorm{x^\top \Phi F - \left[x^\top {P^\pi} \Phi\right]_{\mathrm{sg}}}^2}, \\
    \text{TD Learning: }&& L_{\text{td}}(\Phi,\hat{V}) =& \,\EEX{x \sim \mathcal{D}}{\ltwonorm{x^\top \Phi \hat{V} - \left[x^\top  \left(r^\pi + \gamma P^\pi \Phi \hat{V}\right)\right]_{\mathrm{sg}}}^2}.
\end{align*}

Following the nomenclature of \citet{vaml}, we call the first two losses as \emph{decision-agnostic}, and TD-learning as \emph{decision-aware} as it depends on information specific to decision problem.
To analyze the discrete-time learning dynamics, we study the analogous continuous-time gradient flow, which allows us to use the toolkit of dynamical systems theory.
Writing $\alpha_\Phi$ and $\alpha_F$ for the representation and model learning rates, i.e., the latent self-prediction dynamics are
\begin{align*}
    \ddt{t} \Phi_t &= -\alpha_\Phi\nabla_{\Phi_t} {L}_{\text{lat}}(\Phi_t, F_t) = -2 \alpha_\Phi (\Phi_t F_t - P^\pi \Phi_t)  F_t^\top,\\
    \ddt{t} F_t  &= -\alpha_F\nabla_{F_t} {L}_{\text{lat}}(\Phi_t, F_t) = -2 \alpha_F \Phi_t^\top(\Phi_t F_t - P^\pi \Phi_t).
\end{align*}

We primarily consider the \emph{two-timescale} regime, under the assumption that $\alpha_F\to \infty$ \citep{tang2022understanding}. Intuitively, this describes a learning setup in which the latent model is learned ``much faster'' than the latent mapping. 
This results in the following dynamics for self-predictive learning:
\begin{equation}
    \label{eq:BYOLTwoTimescale}
    F_t^* = \left(\Phi_t^\top\Phi_t\right)^{-1} \Phi_t^\top P^\pi \Phi_t, \quad \frac{d}{dt}\Phi_t = \left(I-\Phi_t\left(\Phi_t^\top \Phi_t\right)^{-1}\Phi_t^\top\right)P^\pi \Phi_t {F_t^*}^\top.
\end{equation}

\section{Formalizing the impact of distractions and observation functions}
\label{sec:formalism}
To bridge the theory-practice gap in feature learning, we formalize two structures found in MDPs found in Deep RL benchmarks that have, to the best of our knowledge, not appeared in work analyzing feature learning: observation functions and distractions.
To allow a close comparison with previous work, our changes to the formalism are minimal on purpose, while still highlighting the important role these changes play in different loss functions.

\subsection{Observation functions}
Previous literature \citep{tang2022understanding,tang2023towards,lelan2022generalization} has eschewed the underlying observation of states in their analysis of representation dynamics. 
The correctness of the dynamical system in \autoref{eq:BYOLTwoTimescale} hinges on the fact that $\mathbb{E}[xx^\top ] = I$, which implies uncorrelated state representations for each state $x$ and a uniform distribution over states.
The simplest form of such a representation would be a \emph{one-hot} vector, a representation for the $i$-th state in which all entries are 0, except the $i$-th, which is 1.

This leads to an assumption that the features for the underlying states can be learnt independently.
With a one-hot representations, the features of $x$ are simply $x^\top \Phi = \Phi[i]$, the $i$-th row of $\Phi$.
A more realistic setting, which we focus on, is considering observation functions acting on the underlying states. 
This allows for representing systems where some states have correlated observations, which may be helpful \emph{or} harmful for the RL problem. 
We provide a motivating example and a more extensive discussion regarding the effect observation functions on the learning process in \autoref{app:observation_motivation}.

We briefly state this for reference later.
\begin{assumption}
    \label{assumption2}
    Let an \emph{observation function} for a finite state space MDP be an invertible matrix $\mathcal{O} \in \mathbb{R}^{n \times n}$.
\end{assumption}

\textbf{Formalization:} To introduce an observation function while remaining in the regime of analyzing linear networks and finite state problems, each one of $n$ states is mapped to a unique $n-$dimensional observation vector by an invertible \emph{observation matrix} $\gO \in \vecR{n}{n}$.\footnote{We could also consider projection into higher dimensional spaces $\gO \in \vecR{n}{d}$ with $d > n$, without violating the Markov assumption, but this leads to additional complications (working with pseudo-inverses instead of inverses) which do not contribute meaningfully to the insights in this work.}
Invertibility is assumed to ensure the Markov property with linear function approximation.

This change from one-hot vectors to arbitrary vectors allows us to account for similarity.
For example, if two states have almost identical observation vectors, they will be mapped to similar points in the latent representation space unless the features directly counteract this.
We study the impact of changing the observations with a linear reparameterization in \autoref{sec:observation} by replacing $x$ with $\bar{x} = \mathcal{O}^T x$.

\subsection{Distracting state dynamics}
In addition to observation functions, another common problem that many reinforcement learning algorithms face are distractions.
While distractions have been a focus of empirical work studying the relative efficacy of different auxiliary tasks \citep{ni2024bridging}, a simple formalism whose effect on learning dynamics can be analyzed has not been presented. 
We propose to model an MDP with distractions using factored MDPs \citep{boutilier2000stochastic}.

\begin{definition}[A factored MDP model of a distraction]\label{def:distracting}
Let $M=(\mathcal{M}, P_{M},R_M)$,  $N=(\mathcal{N},P_N,R_N)$ be a pair of Markov decision processes.
The product process $M \otimes N$ is a MDP with state space $\mathcal{M}\times \mathcal{N}$, transition kernel $P_M\otimes P_N$ (where $\otimes$ signifies the Kronecker product), and reward function $R_M\otimes \vone +\vone \otimes R_N^\top$.

If $R_N = 0$, we refer to $N$ as a distracting process, as it does not contribute to the reward.
\end{definition}

This process models a common occurrence: two non-interacting processes unfold simultaneously, with the states being a combination of the two.
Such a process can model a well-studied form of distraction, the background distractions in  \cite{Stone2021TheDC} or the random observation dimension in \citet{nikishin2021control,voelcker2022value}.
In this case, the foreground process $M$ is assumed to carry the reward information, while the reward vector of the background process $N$ is 0.
We review important properties of the Kronecker product in \autoref{app:linalg}.

Note that our formalizations of observation functions and distracting processes is distinct from the assumptions in \emph{linear MDPs} \citep{jin2020provably}. Concretely we do not assume that the processes are low-rank compressible, just that they are factorizable.

\section{Reconstruction and self-prediction losses}
\label{sec:stand_alone_tasks}

In this section and the next, we present an analysis of the stability conditions of reconstruction and self-prediction losses with linear networks.
Using this analysis, we obtain several \emph{insights}, qualitative predictions about how we expect the studied losses to behave in more complicated scenarios.
These \emph{insights} present the basis for our empirical comparison in \autoref{sec:empirical}.

\subsection{Case 1: Orthogonal state representations}

\newcommand{\hP}[0]{\hat{P}}

\cite{tang2022understanding} show that for symmetric MDPs, latent self-prediction converges to subspaces spanned by eigenvectors of $P^\pi$. 
We extend this result in the following sense: if $P^\pi$ has positive real eigenvalues, invariant sub-spaces which are not spanned by the top-k eigenvectors are unstable for gradient descent.\footnote{The assumption of real \emph{positive} eigenvalues is both more and less restrictive than the symmetry assumption made by \citet{tang2022understanding}. An extension of our result to negative eigenvalues is presented in \autoref{app:proofs}, together with an extended comparison to the results obtained by \citet{tang2022understanding} and \citet{lelan2023bootstrapped}.}
It is interesting to note that the resulting features are identical to those obtained using the multi-reward approach described by \cite{lelan2023bootstrapped} (albeit under slightly different technical conditions), which highlights the close connection between the self-predictive approach and bootstrapped generalized value function learning.
This furthermore suggests that using random rewards as auxiliary objectives \citep{farebrother2023protovalue} could result in very similar features as using self-prediction, which presentes an interesting avnue for further empirical study.

\begin{assumption}
\label{assumption3}
    Assume a two-timescale scenario and $F_0$ being initialized with full rank, and hence the non-collapse property ($\Phi_t^\top\Phi_t = \Phi_0^\top\Phi_0$) \citep{tang2022understanding} holds.
\end{assumption}
When referring to eigenvectors and singular vectors, we mean the \emph{right} vectors of the corresponding matrices unless stated otherwise.
We now present our first theoretical result.

\begin{restatable}[Stationary points of latent self-prediction]{proposition}{BYOLGradientFlow}\label{prop:1}
Assume \autoref{assumption1} and \autoref{assumption2} hold.
Furthermore, suppose $P^\pi$ is real diagonalizable. 
If the columns of $\Phi_t$ span an invariant subspace of $P^\pi$, $\Phi_t$ is a stationary point of the dynamical system.
Furthermore, if $P^\pi$ is real-diagonalizable with positive eigenvalues, all invariant subspaces not spanned by the top-k eigenvectors sorted by eigenvalue are asymptotically unstable for gradient descent.
\end{restatable}

This implies that even without the assumptions of symmetry of $P^\pi$ required by \citet{tang2022understanding}, the dynamics of latent self-prediction will tend to converge to invariant subspaces spanned by eigenvectors with large eigenvalues as other invariant subspaces are unstable.
This is important as we expect these to be more important for representing potential reward functions in the environment \citep{lelan2023bootstrapped}. 

We can contrast this with the features learned by a reconstruction loss.
We write $\mathrm{span}(A)$ for both the span of the column vectors of $A$ or for the span of a set of vectors $A$, depending on context.
\begin{restatable}[Stationary points of reconstruction]{proposition}{ReconstructionStationaryPoints}\label{prop:2}
Assume \autoref{assumption1} and \autoref{assumption2} hold. Write $(u_1,\dots,u_n)$, $(v_1,\dots,v_n)$ for the left and right singular vectors of $P^\pi$ sorted in descending order by singular value. Any stationary point $(\Phi^*, F^*, \Psi^*)$ of $L_\text{rec}$ under the two timescale scenario satisfies $\mathrm{span}\,(\Phi^*)=\mathrm{span}\left(\{u_1,\dots,u_k\}\right)$, $\mathrm{span}\,({\Psi^*}^\top)=\mathrm{span}\left(\{v_1,\dots,v_k\}\right)$.
\end{restatable}

Features of this form have been studied extensively and the convergence properties of linear auto-encoders are well understood \citep{baldi1989neural,pretorius2018learning,bao2020regularized}.

\autoref{prop:1} and \autoref{prop:2} together show that there is a subtle but important difference between latent self-prediction and observation reconstruction: the features will converge to eigenspaces in the former case, and to singular space in the latter case.
Note that if $P^\pi$ is a symmetric matrix, then the singular spaces and the eigen-spaces coincide and latent self-prediction and reconstruction converge to the same features \citep{tang2022understanding}.

\citet{behzadian2019fast} show that top $k$ singular vectors are optimal low-rank linear features when making no assumptions on the reward, meaning observation reconstruction should lead to the best features when considering every possible bounded (or unknown) reward.
\citet{behzadian2019fast} and \citet{lelan2023bootstrapped} both highlight that if eigenvectors and singular vectors differ, singular vectors often lead to better performing features.

\begin{tcolorbox}[boxrule=0.2mm,colback=white,colframe=uoftblue,boxsep=0pt,top=3pt,bottom=5pt]
\begin{insight}[Optimality of observation prediction] The features learned by observation prediction are in general superior to those of latent self-prediction, when using solely one of these as the loss function.
\label{insight1}
\end{insight}
\end{tcolorbox}

\subsection{Case 2: Observation function dependence}
\label{sec:observation}

Recall that the gradient dynamics presented in (\ref{eq:BYOLTwoTimescale}) and analyzed in \autoref{prop:1} and \autoref{prop:2} rely on the assumption that $\mathbb{E}[xx^\top] = I$ (see \autoref{assumption1}).
We now introduce the observation matrix $\gO$, which leads to correlations between different features.
To do this, we simply replace every occurence of $x^\top$ in the losses presented in \autoref{sec:background} with $x^\top \gO$.
We assume that $x$ is a one-hot vector as discussed before and the coverage is still uniform (\autoref{assumption1} still holds), so all correlation between states arise as $\mathbb{E}[\gO^\top x x^\top\gO] = \gO^\top \gO$.

It is important to note that for BYOL and TD this rewriting leads to a linear basis change of $\Phi$ compared to the original loss, as each occurrence of $\gO$ is multiplied by $\Phi$. The only loss for which this is not the case is the reconstruction approach.

\begin{restatable}{proposition}{ReparameterizationInvariance} 
Assume \autoref{assumption1}, \autoref{assumption2}, and \autoref{assumption3} hold. Let $\{\Phi^*_\mathrm{lat/td}\}$ be the set of critical points of $L_\mathrm{lat}$ or $L_\mathrm{td}$ respectively.
Then $\gO^{-1}\Phi^*_\mathrm{lat/td}$ are stationary points for the reparameterized losses $L_\mathrm{lat}^\gO$ and $L_\mathrm{td}^\gO$.\footnote{Due to space constraints, we present the full equations in the proof.} Furthermore, if $\Phi^*_\mathrm{lat/td}$ is an asymptotically stable point of $L_\mathrm{lat/td}$ that has a Jacobian with all negative eigenvalues, $\gO^{-1}\Phi^*_\mathrm{lat/td}$ is an asymptotically stable point of $L_\mathrm{lat/td}^\gO$.
\end{restatable}

Note that while the stationary points and asymptotic stability conditions of the gradient flow might be unaffected by the introduction of observation distortions, the same might not be true for the dynamics of descent with finite step sizes.
The numerical conditioning of the involved matrices change depending on $\gO$ and so the impact of discretization due to finite step sizes changes the resulting dynamical system.

\begin{restatable}{proposition}{ReparameterizationInvarianceObs} 
 Assume \autoref{assumption1}, \autoref{assumption2}, and \autoref{assumption3} hold. Let $(u_1,\dots,u_n)$, $(v_1,\dots,v_n)$ be the left and right singular vectors of $\gO^{-1}P^\pi\gO$. 
Any stationary point $(\Phi^*, F^*, \Psi^*)$ of $L_\text{rec}^\gO$ satisfies $\mathrm{span}\,(\Phi^*)=\mathrm{span}\left(\{u_1,\dots,u_k\}\right)$, $\mathrm{span}\,({\Psi^*}^\top)=\mathrm{span}\left(\{v_1,\dots,v_k\}\right)$.
\end{restatable}

The singular value decomposition of $\gO^{-1}P^\pi\gO$ will in general not have a clearly interpretable relationship to that of $P^\pi$ and $r$, so the optimality result obtained by \citet{behzadian2019fast} do not hold in this case.
However this does not mean that different observation functions will always harm the ability of the reconstruction loss to obtain good features.
Consider for example an observation transformation that maps states directly to value and reward function.
This would clearly be an example of a helpful observation transformation.
However, in general we conjecture that arbitrarily changing the observation function will harm the reconstruction loss approach.
A more detailed analysis involving the reward function of the problem being solved and its connections to the observation model are an exciting avenue for future work.

\begin{tcolorbox}[boxrule=0.2mm,colback=white,colframe=uoftblue,boxsep=0pt,top=3pt,bottom=5pt]
\begin{insight}[Observation dependence of autoencoder models] 
\label{insight2}
Due to the invariance properties of latent self-prediction, we expect the performance of latent self-prediction to suffer less than the performance of observation reconstruction when perturbing the observation space arbitrarily.
\end{insight}
\end{tcolorbox}

\section{Understanding the effects on value function learning}
\label{sec:auxilliary_tasks}

The optimality of a representation for value estimation depends non-trivially on the structure of the reward structure of the MDP. Previous works \citep{behzadian2019fast, bellemare2019geometric, lelan2022generalization} attempted to reason about the optimality of representations without relying on the reward structure, by arguing that certain subspaces (such as the span of top-$k$ eigenvectors or top-$k$ singular vectors) are optimal given reward agnosticism.

Now we take a differing approach by taking the value function structure into account. Furthermore, we argue that the top-$k$ eigenspaces (resp. singular spaces) are not always optimal. Indeed, we demonstrate in \cref{app:distraction_motivation} that with distractions, these subspaces can be particularly poor.

We begin by formalizing the reward function structure we will analyze. 
Let us write $w_1,\dots,w_n$ for the eigenvectors of $P^\pi$. 
We will assume that $r^\pi$ has a low-dimensional structure in the following sense: 

\begin{assumption}\label{ass:low_rank}
    $\exists i_1,\dots,i_m \in \{1,\dots,|\mathcal{X}|\}$ such that $r^\pi \in \mathrm{span}(w_{i_1},\dots,w_{i_m})$, and $m\leq k.$ Let furthermore $\{{w_i}_1,\dots,{w_i}_m\}$ be a minimal basis in the sense that ${w_i}_n {{w_i}_n}^\top r^\pi \neq 0$ for all $n$. 
\end{assumption}
We now write the summed losses $$L_{\text{rec}+\text{td}}(\Phi,F,\psi,\bar{r})= L_\text{rec}(\Phi, F, \psi)+L_\text{td}(\Phi, \bar r)$$ and $$L_{\text{lat}+\text{td}}(\Phi,F,\bar{r})= L_\text{lat}(\Phi, F)+L_\text{td}(\Phi, \bar r).$$

\begin{restatable}{proposition}{BYOLCombined}\label{prop:BYOLCombined}
    Suppose that $P^\pi$ is real diagonalizable, and that \autoref{assumption1}, \autoref{assumption3}, and \autoref{ass:low_rank} hold. 
    There exists a non-trivial critical point $\Phi^*$ of the two-timescale TD loss $L_\text{td}$ such that $\mathrm{span}(r^\pi)\subseteq \mathrm{span}(\Phi^*)$. 
    Furthermore, $\Phi^*$ is a critical point of the two-timescale joint loss $L_{\text{td}+\text{lat}}$. Therefore combining $L_\text{TD}$ and $L_\text{Lat}$ does not exclude the existence of a stationary point with $0$ value function approximation error.
\end{restatable}

We leave the extension of the stability result for TD learning to the joint loss case open for future work.
Note that without the addition of TD learning, the latent loss would stabilize the top-k eigenspace representation, but we hypothesize that this behavior changes when combining the losses.

\begin{restatable}{proposition}{ReconsCombined}
    Let \autoref{assumption1}, \autoref{assumption2}, and \autoref{ass:low_rank} hold. If the reward spanning eigenvectors do not lie within the span of the top-k singular vectors, $\mathrm{span}\,(w_{i_1},\dots,w_{i_m}) \not\subseteq \mathrm{span}\,(u_1,\dots,u_k)$, the critical points of the two-timescale joint loss $L_{\text{td}+\text{rec}}$ are guaranteed to not be minimizers of the value function approximation error. 
\end{restatable}

Contrasting these propositions suggests that when combined with TD learning losses, latent self-prediction can be a more helpful auxiliary task. Indeed, when combining it with TD learning we can still guarantee that there exists an optimal combined solution.
This does not hold for the reconstruction loss, where we can construct cases in which the joint loss leads to worse TD error.

\begin{tcolorbox}[boxrule=0.2mm,colback=white,colframe=uoftblue,boxsep=0pt,top=3pt,bottom=5pt]
\begin{insight}[Latent self-prediction as an auxiliary task] For good performance across a wide variety of tasks, latent self-prediction needs to be combined with TD learning as an auxiliary task. It is a preferable auxiliary task to observation prediction in most scenarios, but especially in scenarios with distracting processes.
\label{insight3}
\end{insight}
\end{tcolorbox}

\section{Empirical study of theoretical results in deep learning based settings}
\label{sec:empirical}

We aim to empirically verify the statements marked as \emph{``Insights''} throughout the paper: superiority of observation prediction as a standalone feature learning loss (\autoref{insight1}), impact of the observation function on the different loss functions (\autoref{insight2}), and the relative strength of latent-self prediction as an auxiliary loss compared to reconstruction (\autoref{insight3}). 
As our theory addresses the simplified setting of policy evaluation with linear models, we seek to test if the insights transfer to the more common setting of control with neural networks.
Across all experiments, we report mean performance over 30 seeds and shaded $95$ bootstrapped confidence interval.

To test these hypotheses, we use the MinAtar suite of five Atari inspired videogames \citep{young19minatar} and the DMC 15 suite  \citep{tunyasuvunakool2020}.\footnote{DMC experiments are presented in the appendix due to space constraints.}
Both are small enough to perform thorough investigations, while providing non-trivial observation spaces and dynamics.
Detailed information about the implementation and hyperparameters can be found in \autoref{app:empirical}.

\begin{figure}[b]
    \centering
    \includegraphics[width=\textwidth]{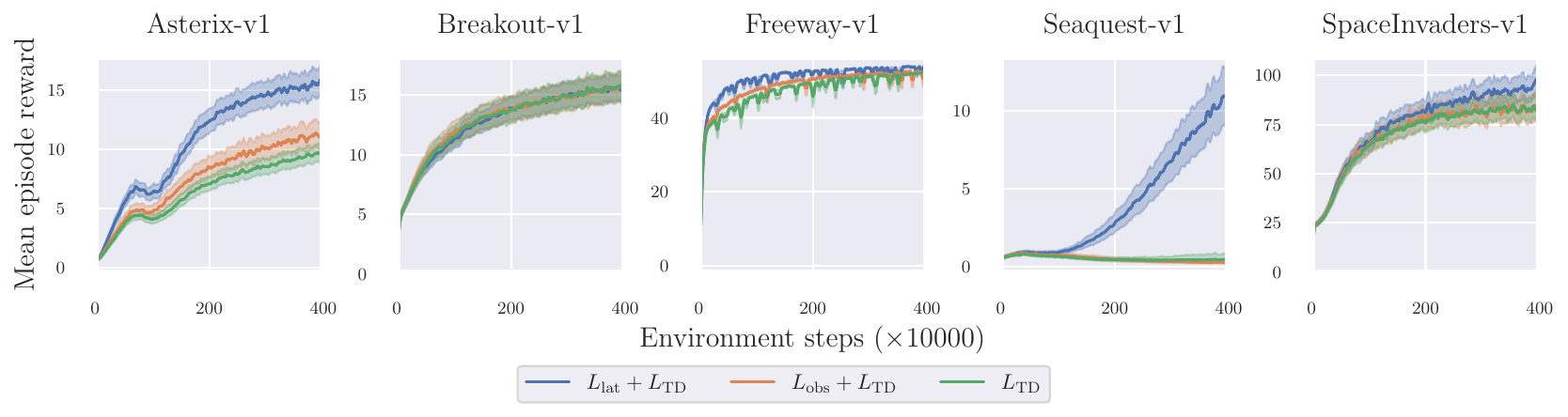}
    \caption{Auxiliary task setup: Performance of all losses on the observation space as given without changes to the environment.}
    \label{fig:aux}
\end{figure}

\begin{figure}[t]
    \centering
    \includegraphics[width=\textwidth]{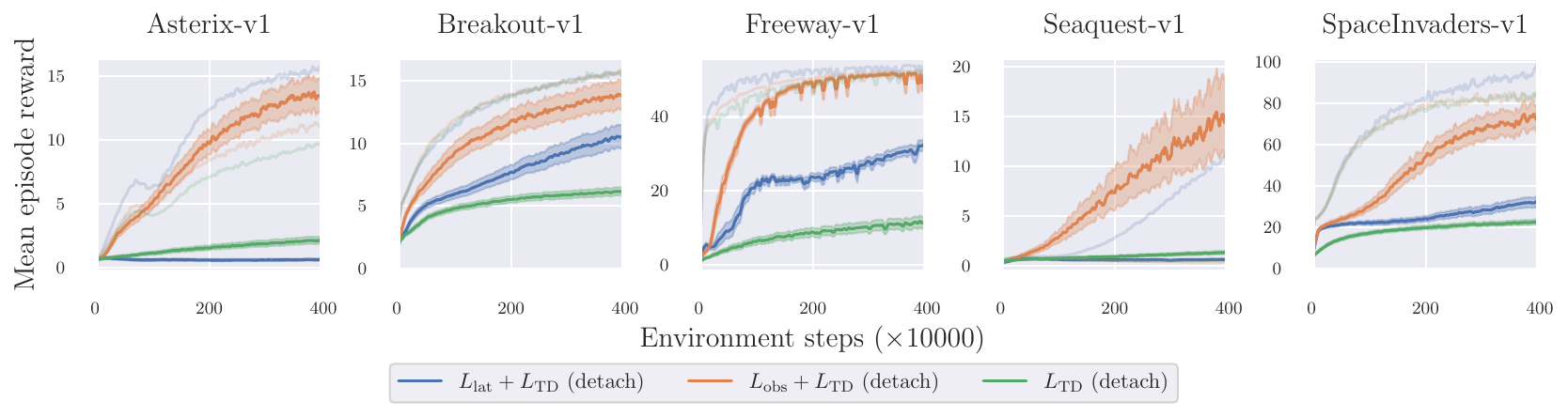}
    \caption{Stand-alone setup: Performance of all losses on the observation space as given without changes to the environment. The DQN baseline is using random features, which are not updated, to verify that learning features is indeed superior to a random feature baseline.}
    \label{fig:sta}
\end{figure}

\textbf{Auxiliary task learning vs general purpose feature learning (\autoref{fig:aux} and \autoref{fig:sta}):}
First, we compare both the auxiliary task and stand-alone feature learning scenarios.
As expected from prior work \citep{jaderberg2016reinforcement,schwarzer2021dataefficient,farebrother2023protovalue}, in all cases using an auxiliary loss performs no worse (and often better) than vanilla DQN.
We find that as expected from \autoref{insight3}, latent self-prediction is a stronger auxiliary loss function than observation reconstruction in three out of five environments.
However, when using the decision-agnostic losses alone, we clearly see observation reconstruction performing significantly better than latent self-prediction, which fails to learn any relevant features in several cases. This verifies \autoref{insight1}.

Curiously, in the case of the Seaquest environment, we find that using observation prediction alone outperforms using it as an auxilliary task strongly, and performs on par with the auxiliary task variant of the latent self prediction loss.
Seaquest also has the sparsest reward structure in the test suite, which can make it a challenging environment for DQN. In this case, features based purely on the observed transition might allow for better policy learning.

This highlights that no algorithm is clearly superior in all settings and the reward and observation structure is very relevant for the performance of each loss.

\begin{figure}[t]
    \centering
    \includegraphics[width=\textwidth]{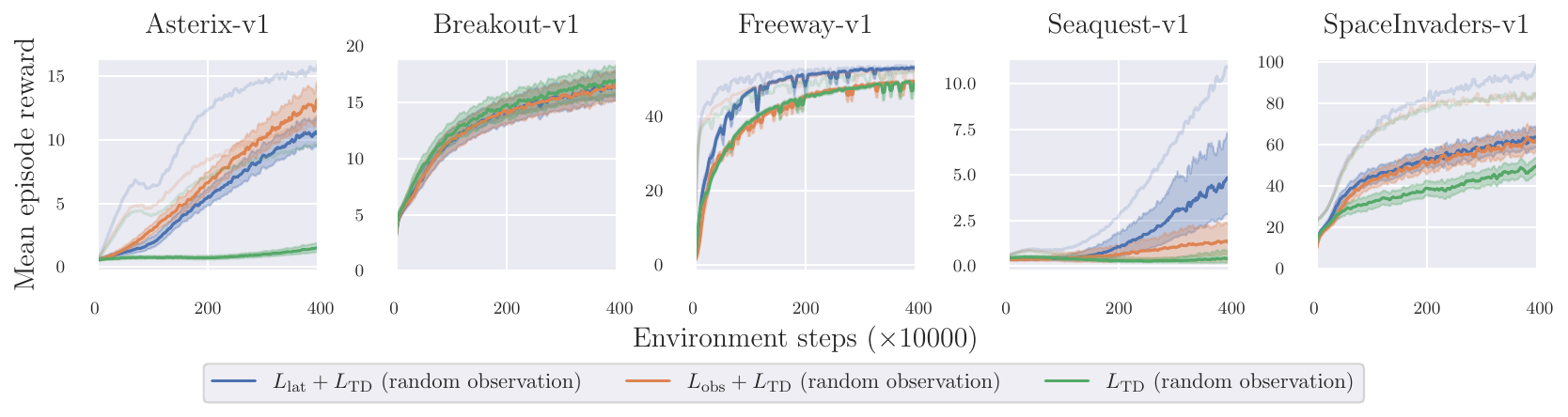}
    \caption{Distorted observation function with a random transformation.}
    \label{fig:dist}
\end{figure}

\textbf{Observation space distortions (\autoref{fig:dist}):} To test the impact of changing the observation function, we sample a random binary matrix and multiply it to the flattened observation vector. We then reshape the observation to the original shape again.

All algorithms show themselves to be strongly impacted by this random observation distortions, which suggests that our claim of invariance of self-prediction relies too strongly on the linear gradient-flow limit.
This can in part be explained by the use of a convolutional layer in the standard baseline implementation of DQN which we adapted.
However, we still find that at least on two environments (Seaquest and Freeway), the latent self-prediction auxiliary task is able to recover more of the original performance than either observation prediction or DQN.
Interestingly, the DQN baseline seems to suffer the most from the introduction of the observation change, which suggests that correlations of the existing observation space play an important role in learning correct value function prediction.

Overall, we find that \autoref{insight2} does not fully translate to the more complex test setting.
In part, this may be explained by the fact that the original observation spaces of the test environments already violate our assumptions for the one-hot encoding.
In addition, introducing linear correlation might not impact non-linear model learning in the same way it would impact linear models.
This highlights the need for more in-depth research on the interplay between given observation space and feature learning.

\begin{figure}[t]
    \centering
    \includegraphics[width=\textwidth]{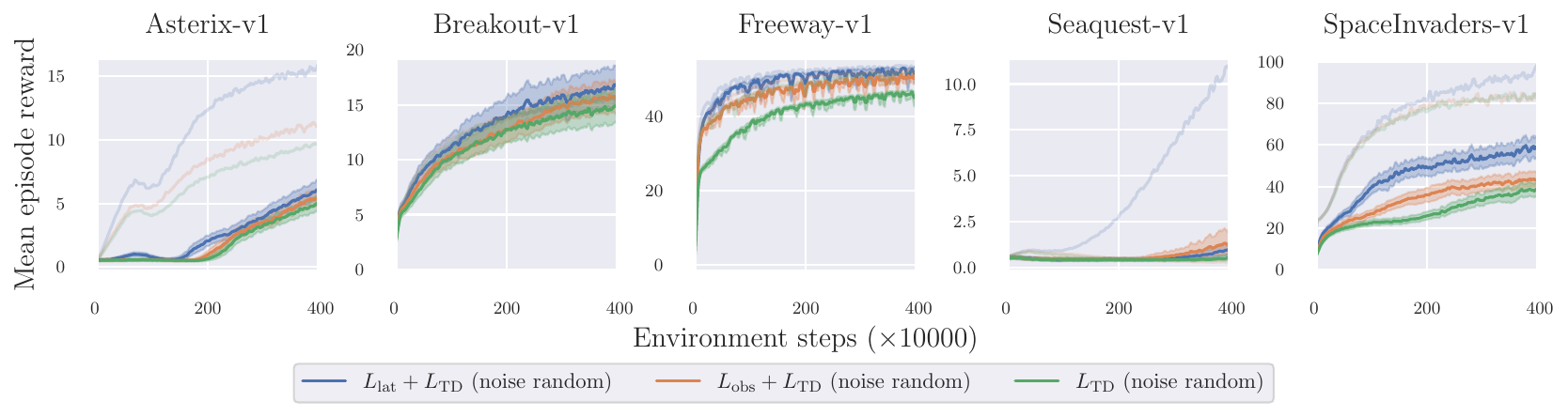}
    \caption{Appending random noise channels.}
    \label{fig:ran}
\end{figure}

\begin{figure}[t]
    \centering
    \includegraphics[width=\textwidth]{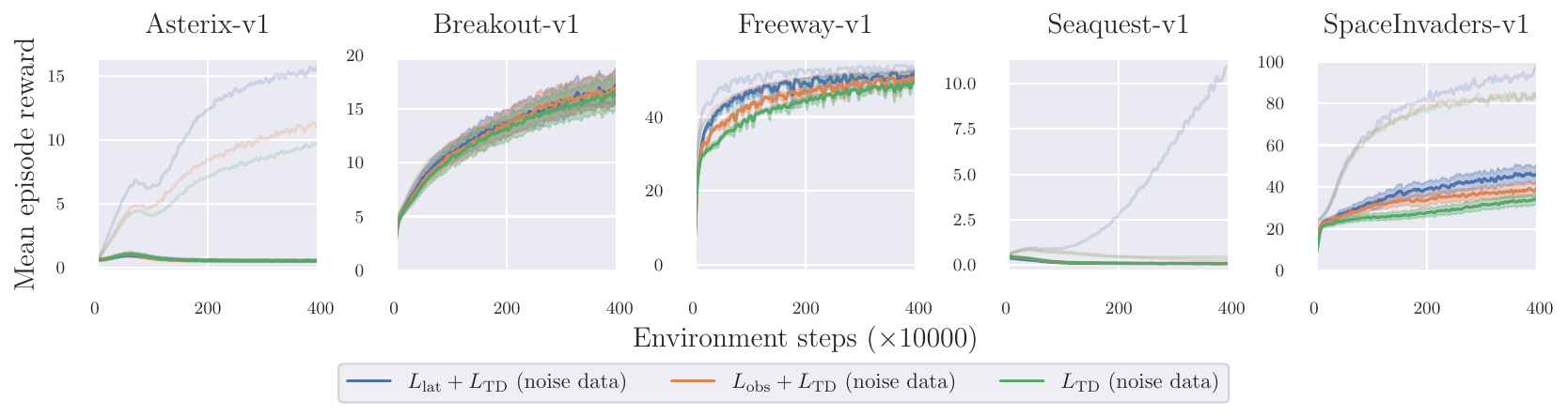}
    \caption{Appending structured noise channels using the Freeway environment.}
    \label{fig:dat}
\end{figure}

\textbf{Distractions (\autoref{fig:ran} and \autoref{fig:dat}):}
As our results are dependent on the spectral structure of the environment, different distraction models can be assumed to have differing impact on the efficacy of the tested losses.
This behavior is dependent on the structure of the noise.
If the distraction does not strongly change the top-$k$ singular or eigenspaces, it will be less problematic for the auxiliary tasks, especially for observation reconstruction.
Testing the impact of different distraction models on the top-$k$ spaces is out of scope for this work, but we conjecture that fully random noise has less structure than distractions following clear patterns.

Therefore, we consider two simple distraction models in our experiments. 
The distractions are concatenated to the original observations along the channel dimension.
First, we use random noise sampled independently for each state from a Bernoulli distribution.
As there is no predictable structure in this noise, we expect all algorithms to be able to deal with this distraction better.
Second, we choose one of the environments at random (Freeway-v1) and concatenate two copies to the observation space of each environment.
The dynamics are obtained by sampling a random action at each timestep independent from the policy and stepping the distraction environment with it.

We find a small advantage in some environments to using the latent self-prediction loss and using random noise, and no clear advantage from any algorithm in the structured noise case.
Structured noise poses a much larger challenge to most algorithms, completely preventing learning in several cases.
This partially validates that not only the presence or absence of noise matters, but also how it changes relevant quantities, e.g. the eigenvalues of the transition kernel.

In the continuous control experiments presented in \autoref{app:mujoco_results}, we find that the self-prediction loss performs generally better than in the MinAtar suite.
As the observation and reward structure differs between these two benchmark suite, this obvservation lends more credibility to our claims that the observation structure impacts the performance of algorithms.

\section{Conclusions}

When choosing an RL approach to use, practitioners are overwhelmed with a variety of loss function choices, without clear indication which one will be preferable in what scenario.
In our work, we introduce analytically tractable notions of distractions and observation functions.
With these we predict the performance of latent self-prediction and observation reconstruction as stand-along feature learning methods and auxiliary tasks, and study our theoretical insights empirically.
Our evaluation lends credibility to the use of simple surrogate models to obtain practically relevant insights into algorithmic performance.
However, in several cases we also find deviations between our predictions and more complex benchmarks.
Therefore, while we claim that our experiments have the ability to guide the choice of algorithms for applied settings, there is still a sizeable gap between theory and practice that remains to be bridged in future work.

We also note that our experiments showed substantial differences in behavior of auxiliary losses both within and across benchmarks and different noise distractions.
Previous work that studied the effect of distraction \citep{nikishin2021control,voelcker2022value,ni2024bridging} did not discuss their distraction model in further details. 
In light of our results, we suggest that empirical research should be careful about the choice of benchmark and experimental setup and discuss the implications of the empirical setup explicitely.

One of the most important gaps between the work presented in this paper and the behavior of online algorithms is the restrictive assumption of the fixed policy evaluation case.
Therefore one of the most exciting avenues for future work is analysing policy improvement, where the underlying dynamics of the environment change due to the policy updates.
On the empirical side the surprising effectiveness of the observation prediction loss on the Seaquest environment highlights the fact that even within benchmark suites, differences in the reward functions and observation models can lead to differing rankings between algorithms.
This further highlights the necessity of studying the structure of MDPs and to design algorithms that are robust to different structures, or adapt to them automatically.

\subsubsection*{Acknowledgments}
\label{sec:ack}

We would like to thank the members of the Adaptive Agents Lab and the Toronto Intelligent Systems Lab for their help and feedback on the ideas and the draft.
CV thanks Charline Le Lan and Anvith Thudi for help with the stability proofs and other mathematical details.
AMF acknowledges the funding from the Canada CIFAR AI Chairs program, as well as the support of the Natural Sciences and Engineering Research Council of Canada (NSERC) through the Discovery Grant program.
Resources used in preparing this research were provided, in part, by the Province of Ontario, the Government of Canada through CIFAR, and companies sponsoring the Vector Institute.

\bibliographystyle{rlc}
\bibliography{bibliography}

\newpage
\appendix

\section{Motivating examples}
\subsection{Motivating example for observation spaces}
\label{app:observation_motivation}

To further motivate our focus on observation models, consider the common problem of representing rotational angles.
As these are continuous values, discretized one-hot representations may introduce errors that may harm efficient control.
When choosing a continuous representation, designers are faced with the choice between representing the angle as $\omega\in[-\pi,\pi)$ (the exclusion of the right endpoint is arbitrary), or by a decomposition into $[\sin(\omega),\cos(\omega)]\in [0,1]^2$.
The choice of representation has measurable impact on the ease of learning: the former is 1D instead of 2D, which can reduce the size of networks needed when dealing with many angles. However a complication with the first representation is it is discontinuous at the right endpoint, in the sense that $\lim_{x\to \pi}x = -\pi$ (this is generally due to the structure of $\mathbb{R}/2\pi\mathbb{Z}$). This creates a peculiar continuity condition for functions on this representation space: for a function $f$ to be continuous, it must be continuous and also satisfy the boundary condition $\lim_{x\to \pi}f(x) = f(-\pi)$. Without explicitly enforcing this, the majority of functions learned on this domain (such as estimated value functions) will be discontinous, leading to additional difficulties in the learning process. On the other hand, this issue does not exist for the second representation, potentially making learning much easier.

The important question is that of \emph{similarity} and \emph{continuity}: in our pendulum example, the more compact representation breaks the intuitive notion that states that behave similariy should be mapped to close representations.

\subsection{Distractions}
\label{app:distraction_motivation}

The optimality of features can be measured in how close the projection of the true value function onto their span is to the ground truth, i.e. in the $L_2$ norm.
This raises the question under what conditions the top $k$ eigenvectors or singular vectors would not span the value function well.
For this, we turn to our notion of an MDP with distractions.

\begin{restatable}[Suboptimality of top $k$ eigenspaces with distractions]{proposition}{topkDistracted} \label{prop:SubptimalTopKProduct} Assume an MDP with distraction composed of two independent processes $\gM$ and $\gN$ according to \cref{def:distracting}.
Let $v_1,\dots,v_n$ and $u_1,\dots,u_m$ be the eigenvectors of $\gM$ and $\gN$ respectively, with associated real eigenvalues $\lambda_j$ and $\mu_i$ ordered.
Assume $\forall i < k: \mu_i > \lambda_2$ and $r_\gN = 0$.
Let $U_k$ be an orthonormal basis for the top-k eigenspace of $\gM \otimes \gN$.
Then $\mathrm{span}(U_k) = \mathrm{span}(\{\vone \otimes v_i|\forall i \leq k\})$ and $\mathrm{proj}_{U_k}(r_\gM \otimes \vone) = (\sum_{i=0}^m r_i / n) \vone_{n\cdot m}$.
\end{restatable}

This means that there is a natural notion of a process in which the top-$k$ eigenvectors do not span the reward function well.
In this case, the distracting process has larger eigenvalues than the reward-relevant process.
As the eigenvector basis is composed of Kronecker products of the individual eigenvectors, the top-$k$ eigenspace contains redundant copies of the reward relevant processes eigenvectors.
By \autoref{lem:spectrum_rew_value}, this directly implies suboptimal value function approximation.
Note that our assumptions here are restrictive as we consider a \emph{worst case} distraction for clarity, but the problem emerges whenever the distracting process has several large eigenvalues.

\section{Related work}

In this work, we take inspiration from the work of \citet{tang2022understanding} and \citet{lelan2023bootstrapped}, which present analyses of representation dynamics under various losses.
The main purpose of our work is to take a closer look at the implicit assumptions on the MDP structure underlying previous work to obtain a better picture of the empirical behavior of the analyzed algorithms.

\citet{bellemare2019geometric} shows how the space of linear features relates to optimal value functions.
Building on this, \citet{lelan2022generalization} analyzes the optimality of, while \citet{le2021metrics} focus on the topological properties of representation functions.
Our work expands on these by focusing on the properties of features learned by common approaches instead of reasoning about optimal features.

\cite{lyle2021effect} and \cite{farebrother2023protovalue} analyse the impact of random cumulants and auxiliary value function prediction on the loss dynamics.
As \citet{lelan2023bootstrapped} show, the induced gradients are similar to those we analyze for latent self-prediction, adding another avenue for understanding the design of auxiliary tasks.
As latent self-prediction and auxiliary tasks lead to similar stationary points, it is an exciting avenue for future work to use the spectral properties of existing reward functions to design better auxiliary rewards, i.e. in environments with distractions.

\citet{tomar2023learning} asks a closely related question to ours ("What matters for reinforcement learning from pixels?") and provide an extensive empirical study. Many of their findings are corroborated in our experiments and strengthened by our theoretical analysis.
They finding only limited benefit for latent self-prediction, which is different from our finding, however they focused on more complex visual observation spaces, which might highlight that the benefits of this approach are more strongly observed in smaller representation spaces. 

Linear features have also been studied in the context of the graph Laplacian \citep{petrik2007analysis,mahadevan2009learning,wu2018laplacian} and successor features \citet{dayan1993improving,barreto2017successor}, which result in feature construction methods that closely resemble those learned by current methods.
Our work directly relates modern deep learning methods to these classic approaches for feature construction.

Finally, \citet{ni2024bridging} recently published an analysis on the relative performance of different auxiliary tasks in deep reinforcement learning.
Our analysis is orthogonal and complimentary to theirs.
While they focus on highlighting the benefits of latent self-prediction in the idealized limit of perfect predictions, we analyse the case where trade-offs have to be made due to compression.
In addition, while they highlight the noise robustness of latent self-prediction, they show no formalization or theoretical justification.
Our analysis on the other hand shows the strong impact that the form of the distracting dimensions have on the problem solution.

\section{Helpful definitions and lemmata}

This section contains helpful lemmata that are used for our proofs.
Where we took these from existing work, we provide references, otherwise the proofs are our own, although probably also known in the literature.

\subsection{Linear Algebra}
\label{app:linalg}

As before, for matrices $V$ we write $\mathrm{span}(V)$ to denote the span of their column vectors. 

\begin{definition}[Top-$k$ singular vectors]\label{def:topk-single}
Let $P^\pi = U^\top \Sigma V$ be the singular value decomposition, and assume that the diagonal of $\Sigma$ is arranged in decreasing order.
The first $k$ rows of $U^\top$ and the first $k$ columns of $V$ are called the top-k left and right singular vectors, respectively.
\end{definition}

\begin{lemma}[Spectrum of Kronecker product matrix]\label{lem:spectrum_kronecker}
    Consider two non-singular matrices $M$ and $N$. Let $\lambda_i$ be the eigenvalues of $M$ and $\mu_j$ be the eigenvalues of $N$, with eigenvectors $u_i$ and $v_j$ respectively. Then the eigenvalues of $M\otimes N$ are $\lambda_i\mu_j$ with eigenvectors $u_i \otimes v_j$ respectively.
\end{lemma}

\begin{proof}
For any eigenvector $u_i$ of $M$ and $v_j$ of $N$, we have
    $$\left(M\otimes N\right) \left(u_i \otimes v_j\right) = \left(M u_i\right) \otimes \left(N v_j\right) = \left(\lambda_i u_i\right)\otimes\left(\mu_j v_j\right) = \lambda_i\mu_j \left(u_i \otimes v_j\right).$$

    So $u_i\otimes v_j$ is an eigenvector of $M\otimes N$ with eigenvalue $\lambda_i \mu_j$
\end{proof}

\begin{lemma}[Orthogonalized bases and the Kronecker product]\label{lem:orth_kronecker}
    Let $V \in \vecR{n}{m}$ be a matrix. Let $\mathrm{orth}(V)$ be a matrix of any orthonormal basis vectors for the column vectors of $V$. Then $\mathrm{span}(V\otimes \vone) = \mathrm{span}(\mathrm{orth}(V) \otimes \vone)$. Furthermore $(1/k) \mathrm{orth}(V) \otimes \vone$ is an orthonormal basis of $\mathrm{span}(V \otimes \vone)$.
\end{lemma}

\begin{proof}
    Let $V_i \otimes \vone^k$ be any column vector of $V\otimes \vone^k$. Let $\alpha_1,\dots,\alpha_m$ be coefficients so that $V_i = \sum_{j=1}^m \alpha_j \mathrm{orth}(V)_j$. Then $$\sum_{j=1}^m \alpha_j \left(\mathrm{orth}(V)_j \otimes \vone^k\right) = \left(\sum_{j=1}^m \alpha_j \mathrm{orth}(V)_j\right) \otimes \vone^k = V_i \otimes \vone^k,$$ following the standard associative and distributive properties of the Kronecker product.

    As every vector in the original span can be represented in the orthogonalized span, the two are equivalent.

    Finally note that $$\frac{1}{k}\left(\mathrm{orth}(V) \otimes \vone\right)_i^\top \left(\mathrm{orth}(V) \otimes \vone\right)_i = \frac{1}{k}\sum_{j=1}^k \left(\mathrm{orth}(V)_i^\top \mathrm{orth}(V)_i^\top\right) = 1,\,\text{and}$$ $$\frac{1}{k}\left(\mathrm{orth}(V) \otimes \vone\right)_j^\top \left(\mathrm{orth}(V) \otimes \vone\right)_i = \frac{1}{k}\sum_{t=1}^k \left(\mathrm{orth}(V)_j^\top \mathrm{orth}(V)_i^\top\right) = 0.$$
\end{proof}

\begin{lemma}[Reduced rank regression]\label{lem:rrr}
    Let $C,D \in \vecR{n}{n}$, $A\in\vecR{n}{k}$, and $B\in\vecR{k}{n}$ be full rank matrices with $n \geq k$. Let $A^*, B^* = \min_{A,B} \lVert CAB - DC\rVert^2_F$. Let ${u_1,\dots,u_k}$ and ${v_1\dots,v_k}$ be the top-k singular vectors of $C^{-1}DC$ according to \autoref{def:topk-single}. Then $\mathrm{span}(A^*) = \mathrm{span}({u_1,\dots,u_k})$ and $\mathrm{span}(B) = \mathrm{span}({v_1,\dots,v_k})$.
\end{lemma}

This is a reduced rank regression problem \citep{izenman1975reduced} or low-rank matrix approximation problem. 

The unconstrained solution to the problem is given by $\hat{A}\hat{B} = C^{-1}DC$. From the Eckhart-Young theorem, we know that the optimal low-rank approximation to $C^{-1}DC$ is given by the top-k singular vectors.
Therefore $A$ and $B$ span top-$k$ left and right singular vectors respectively.
For a more extensive proof, please refer to \citet{izenman1975reduced}.

\subsection{Stochastic matrices}
\begin{lemma}[Spectrum of a resolvent matrix]\label{lem:spectrum_resolvent}
    Let $A$ be an invertible matrix with unique real eigenvalues and $-1 \leq \lambda_{\min} \leq \lambda_{\max} \leq 1 $. Let $\gamma\in(0,1)$.
    The matrices $A$ and $(I - \gamma A)^{-1}$ have the same eigenvectors and the ordering of the corresponding eigenvalues remains the same.
\end{lemma}

\begin{proof}
    Let $e$ be an eigenvector of $A$ and $\lambda$ the corresponding eigenvalue. Then $$(I - \gamma A)^{-1}e = \sum_{i=0}^n \gamma^n A^ne = \sum_{i=0}^n (\gamma\lambda)^n e = \frac{1}{1 - \gamma\lambda}e.$$ As $(1-\gamma\lambda)^{-1}$ is a monotonic function for $-1 \leq \lambda \leq 1$ and $\gamma\in[0,1]$, the ordering of the eigenvalues remains the same.
\end{proof}

\begin{lemma}[Basis equivalence of linear reward and value function]\label{lem:spectrum_rew_value}
    Let $r$ be the vector representation of the reward function and $V$ of the value function respectively for an MDP with fixed policy and transition matrix $P^\pi$.
    Let $U = \{u_1,\dots,u_n\}$ be the set of eigenvectors of $P^\pi$, and let $U^r \subseteq U$ be a minimal set of eigenvectors so that $r \in \mathrm{span}(U^r)$.
    Then $V \in \mathrm{span}(U^r)$.
\end{lemma}
\begin{proof}
    Let $r = \sum_{i=1}^k \alpha_i\, u^r_i$ be the reward representation in the basis $U^r$. Then, by \autoref{lem:spectrum_resolvent} $$V^\pi = (I - \gamma P^\pi)^{-1} r = \sum_{i=1}^k \frac{\alpha}{1 - \lambda_i\gamma}  u^r_i.$$
\end{proof}

\subsection{Ordinary differential equations}

For describing the stability of ODEs, we use the notion of \emph{asymptotic stability}, with the condition $\Re(\lambda_i) < 0$ for all eigenvalues $\lambda_i$ of the Jacobian at a critical point.

\begin{lemma}[Linear reparameterization of an autonomous ODE]
\label{lem:stability}
Let $y' = f(y)$ be an autonomous ordinary differential equation. Let $y^*$ be any critical point for which $f(y^*) = 0$. Let furthermore $x' = f(A x)$ be a reparameterized autonomous ODE for any invertible matrix $A$.
Then the $x^*$ is a critical point with $f(Ax) = 0$ iff $x^* = A^{-1}y^*$.
Furthermore, the eigenvalues of the Jacobian of $y'$ at $y^*$ are equal to the eigenvalues of the Jacobian of $x'$ at $x^* = A^{-1}y^*$.
\end{lemma}

\begin{proof}
The direction $x = A^{-1}y^* \Rightarrow f(Ax) = 0$ is clear by direct evaluation. We now focus on the direction $f(Ax) = 0 \Rightarrow x = A^{-1}y^*$. Assume that $f(Ax)$ is $0$ and $x = A^{-1}y$ for a $y$ which is not a point satisfying $f(y)=0$. But then $0 = f(AA^{-1}y) = f(y) \neq 0$, a contradiction.

For stability, note that
\begin{alignat*}{2}
    &\qquad&\ddt{t} y & = f(y) \\
    \implies&&\ddt{t} x & = \ddt{y} x \ddt{t} y\\
    && &= A^{-\top} f(y)\\
    \implies&& \ddt{x} \ddt{t} x &= A^{-\top} \ddt{x} f(A x)\\
    && &= A^{-\top}  \ddt{y} f(y) \ddt{x} A x\\
    && &= A^{-\top}  \ddt{y} f(y) A^\top.
\end{alignat*}
This shows that the Jacobian $\ddt{x} f(Ax)|_{x=A^{-1}y_0}$ is similar to the Jacobian $\ddt{y} f(y)|_{y=y_0}$, which means their spectra are identical.
\end{proof}

\subsection{MDP representation and TD learning}

\begin{lemma}[Lemma 5 of \cite{tang2022understanding}]\label{prop:TangResult2}
    Suppose $P^\pi$ is real-diagonalizable, and write $u_1,\dots, u_n$ for its eigenvectors. Then any orthonormal matrix $\Phi$ which has the same span as a set of $k$ eigenvectors is a minimizer of $L_{\text{lat}}$.
\end{lemma}

The next three statements are taken from \citet{ghosh2020representations}.  They address the TD loss wrt to the learned weights $\hat{V}$ and fixed $\Phi$ (compare \autoref{sec:background}).
We changed the notation of the statements to fit our notation here, we have denoted the diagonal matrix of the state distribution as $D$ and assume that $D=I$ in \autoref{assumption1}, while \citet{ghosh2020representations} uses $\Xi$. They use $\theta$ for the value function weights while we use $\hat{V}$.
They also uses $\text{Spec}(A)$ to denote the spectrum, the set of all eigenvalues of a matrix $A$.

The notion of stability used in \citet{ghosh2020representations} is that of convergence to the unique fixed point of the projected Bellman update of the linear ODE induced by the $L_\text{TD}$ loss when fixing $\Phi$. In the two-timescale scenario considered in this paper, this corresponds to the ``inner'' ODE over $\hat{V}$.

\begin{lemma}[Proposition 3.1 of \citet{ghosh2020representations}]\label{prop:gosh1}
    TD(0) is stable if and only if the eigenvalues of the implied iteration matrix $A_\Phi = \Phi^\top D (I - \gamma P^\pi) \Phi$ have positive real components, that is $$\text{Spec}\left(A_\Phi\right) \subset \mathbb{C}_+ \coloneq \{z : \text{Re}(z) > 0 \}.$$
    We say that a particular choice of representation $\Phi$ is stable for $(P^\pi, \gamma, D)$ when $A_\Phi$ satisfies the above condition.
\end{lemma}

\begin{lemma}[Proposition 3.2 of \citet{ghosh2020representations}]\label{prop:gosh2}
    An orthogonal representation $\Phi$ is stable if and only if the real part of the eigenvaluse of the induced transition matrix $\Pi P^\pi \Pi$ where $\Pi = \Phi\Phi^\top$ is bounded above, according to $$\text{Spec}\left(\Pi P^\pi \Pi\right) \subset \{z \in \mathbb{C}: \text{Re}(z) < \frac{1}{\gamma} \}.$$
    In particular, $\Phi$ is stable if $\rho(\Pi P^\pi\Pi) < \frac{1}{\gamma}$.
\end{lemma}

\begin{lemma}[Theorem 4.1 of \citet{ghosh2020representations}]\label{prop:gosh3}
    An orthogonal invariant representation $\Phi$ (meaning $\text{span}(P^\pi \Phi) \subseteq \text{span}(\Phi)$) satisfies $$\text{Spec}\left(\Pi P^\pi \Pi\right) \subseteq \text{Spec}(P^\pi) \cup \{0\}$$ and is therefore stable.
\end{lemma}

As a corollary of their proof we have that 

\begin{lemma}[Corollary of \citet{ghosh2020representations}]\label{gosh:corr}
    Let $\Phi$ be an orthogonal (but not necessarily square) invariant representation of $P^\pi$. Then the spectral radius $\rho(\Phi^\top P^\pi \Phi) \leq 1.$ 
\end{lemma}

The proof follows directly from \autoref{prop:gosh3} by the cyclicality of the spectrum.

The following two results are our own, although closely related results exist in the literature.

\begin{lemma}[Lossless approximation of $V$]\label{lem:lossless_approx}
    Let $\Phi$ be an orthonormal basis of an invariant subspace of $P^\pi$ and let \autoref{ass:low_rank} hold. Let $V = (I - \gamma P^\pi)^{-1} r$ be the value function of $P^\pi$ and $r^\pi$. Then $$\Phi (I - \gamma \Phi^\top P^\pi \Phi)^{-1} \Phi^\top r^\pi = V.$$
\end{lemma}
\begin{proof}
    By \autoref{ass:low_rank} we can find a matrix $A$ so that $r^\pi = \Phi A$ and $\Phi\Phi^\top r^\pi = r^\pi$, and by the invariant subspace assumption, we can find a matrix $B$ so that $P^\pi \Phi = \Phi B$.
    Writing the inverted matrix as an infinite sum, which is valid as the spectrum of $\Phi^\top P^\pi \Phi$ is bounded by 1 following from \autoref{gosh:corr} and Carl Neumann's theorem over power series, we obtain
    \begin{align*}
        \Phi (I - \gamma \Phi^\top P^\pi \Phi)^{-1} \Phi r^\pi &= \Phi \sum_{n=0}^\infty \gamma^n (\Phi^\top P^\pi \Phi)^n \Phi r^\pi \\
        &= \Phi \sum_{n=0}^\infty \gamma^n B^n \Phi^\top r^\pi &\quad (P^\pi \Phi = \Phi B)~\text{and}~(\Phi^\top \Phi = I)\\
        &= \sum_{n=0}^\infty \gamma^n \Phi B^n \Phi^\top r^\pi \\
        &= \sum_{n=0}^\infty \gamma^n {P^\pi}^n \Phi\Phi^\top r^\pi &\quad (\Phi B = P^\pi \Phi)~\text{iterated} \\
        &= \sum_{n=0}^\infty \gamma^n {P^\pi}^n r^\pi &\quad (\Phi\Phi^\top r^\pi = r^\pi) \\
        &= V
    \end{align*}
\end{proof}

\begin{lemma}[Critical points of $L_\text{TD}$]\label{prop:td_critical}
Let \autoref{assumption1}, \autoref{assumption3}, and \autoref{ass:low_rank} hold.
    Assume $\Phi^* \in \mathbb{R}^{n \times k}$ is an orthonormal invariant representation for $P^\pi$ in the sense that ${\Phi^*}^\top \Phi^* = I$ and $\text{span}(\Phi^*) = \text{span}(P^\pi \Phi^*)$. Let furthermore  $r^\pi \in \text{span}(\Phi^*)$ and let $V$ be the corresponding value function. Then $\Phi^*$ is a critical point of $L_\text{TD}$ and for the corresponding weights $\hat{V}^*$ we have $\Phi^*\hat{V}^* = V$.
\end{lemma}

\begin{proof}
    By \autoref{prop:gosh1}, \autoref{prop:gosh2}, and \autoref{prop:gosh3} and the stated assumptions the weights $\hat{V}$ converge. Therefore, we can analyze the induced dynamical system with $\hat{V}*$.

    Find $\hat{V}^*$ as the stationary point of $\nabla_{\hat{V}} L_\text{TD}$
    \begin{align*}
        & \nabla_{\hat{V}} \ltwonorm{\Phi^* \hat{V} - \left[r^\pi + \gamma P\Phi^* \hat{V}\right]_\text{sg}}^2\bigg|_{\hat{V} = \hat{V}^*} = {\Phi^*}^\top\left(\Phi^* \hat{V}^* - \left[r^\pi + \gamma P\Phi^* \hat{V}^*\right]\right) = 0 \\
        \Leftrightarrow \quad & {\Phi^*}^\top \Phi^* \hat{V}^* - {\Phi^*}^T r^\pi - \gamma {\Phi^*}^\top  \Phi^* B \hat{V}^* = ({\Phi^*}^\top \Phi^* - \gamma {\Phi^*}^\top P^\pi \Phi^*) \hat{V}^* - {\Phi^*}^\top r^\pi = 0\\
        \Leftrightarrow \quad & \hat{V}^* = ({\Phi^*}^\top \Phi^* - \gamma {\Phi^*}^\top P^\pi \Phi^*)^{-1} {\Phi^*}^\top r^\pi = (I - \gamma {\Phi^*}^\top P^\pi \Phi^*)^{-1} {\Phi^*}^\top r^\pi
    \end{align*}
    The invertibility of $(I - \gamma \Phi^\top P^\pi \Phi) = \Phi^T D (I - \gamma P^\pi) \Phi = A_\Phi$ is guaranteed as all eigenvectors are nonzero.

    We now show that $\Phi^*$ is a stationary point by showing that $$\nabla_\Phi L_\text{TD}|_{\Phi = \Phi^*} = 0.$$ We note that as $\Phi^*$ spans an invariant subspace of $P^\pi$ there exists an invertible matrix $B$ so that $P^\pi \Phi^*  \Phi^* B$. Therefore $(I - \gamma {\Phi^*}^\top P^\pi \Phi^*) = (I - \gamma B)$.

    \begin{align*}
        \nabla_\Phi L_\text{TD}|_{\Phi = \Phi^*} & =  \nabla_{\Phi} \ltwonorm{\Phi \hat{V}^* - \left[r^\pi + \gamma P\Phi \hat{V}^*\right]_\text{sg}}^2\Bigg|_{\Phi = \Phi^*}\\
        & = \left(\Phi \hat{V}^* - r^\pi - \gamma P\Phi {\hat{V}}^*\right) (\hat{V}^*)^\top\Bigg|_{\Phi = \Phi^*} \\
        & = \left(\Phi^* (I - \gamma B) {\hat{V}}^* - r^\pi \right)(\hat{V}^*)^\top \quad\quad \text{(substitute first occurrence of)~$\hat{V}^*$}\\
        & = \left(\Phi^* {\Phi^*}^\top r^\pi - r^\pi \right)({\hat{V}}^*)^\top\\
        & = \underbrace{\left(r^\pi - r^\pi \right)}_{=0}({\hat{V}}^*)^\top = 0
    \end{align*}
    The final line is due to the fact that the columns of $\Phi^*$ are orthonormal, which means that $\Phi^*{\Phi^*}^\top$ is an orthogonal projection onto the span of $\Phi^*$. To verify, note that $$\left(\Phi^*{\Phi^*}^\top\right)^2 = \left(\Phi^* \underbrace{{\Phi^*}^\top \Phi^*}_{=I}{\Phi^*}^\top\right) = \left(\Phi^*{\Phi^*}^\top\right).$$ Furthermore, by \autoref{ass:low_rank}, $r^\pi \in \text{span}(\Phi^*)$, which means $\Phi^*{\Phi^*}^\top r^\pi = r^\pi$ for an orthogonal projection $\Phi^*{\Phi^*}^\top$.
    Moreover, by \autoref{lem:lossless_approx}, $\Phi^* \hat{V}^* = V$, which concludes the proof.
\end{proof}

This proof closely follows related statements by \cite{ghosh2020representations}, \cite{tang2022understanding}, and \cite{lelan2022generalization}. We repeated the argument here for easier legibility with all assumptions necessary for our work.

\section{Proofs of main results}
\label{app:proofs}

\subsection*{Proofs for Section 4}

\BYOLGradientFlow*
\begin{proof}
    At any stationary point the gradient $\frac{d}{dt}\Phi_t$ must be equal to $0$, which from \cref{eq:BYOLTwoTimescale} means that we must have $\left(\Phi_t\left(\Phi^\top_t \Phi\right)^{-1}\Phi_t^\top - I\right)P^\pi\Phi_t \Phi_t^\top {P^\pi}^\top \Phi_t\left(\Phi^\top_t \Phi\right)^{-\top}=0$.

    Assume that the column vectors of $\Phi^*$ spans an invariant subspace of $P^\pi$. This implies that there exists a full rank matrix $A$ so that $P^\pi \Phi^* = \Phi^*A$.
    Then 
    $$\left(\Phi^*\left({\Phi^*}^\top \Phi^*\right)^{-1}{\Phi^*}^\top - I\right)P^\pi\Phi^* F^*=\Big(\Phi^*\underbrace{\left({\Phi^*}^\top \Phi^*\right)^{-1}{\Phi^*}^\top\Phi^*}_{=I} - \Phi^*\Big)A F^*=0.$$

This proves the first part of the proposition.

There are additional critical points of the differential equation, as discussed by \citet{tang2022understanding}.
In the analysis of stability, we first show the case of critical points corresponding to the claim in the proposition.
We then briefly discuss other cases after the proof.

\textbf{Case 1: $\Phi_t$ spans an invariant subspace of $P^\pi$}
Invariant subspaces correspond to subspaces spanned by right eigenvectors of $P^\pi$.

We write $P$ for $P^\pi$ to reduce notational clutter.
Let $e_1,\dots,e_k$ be the eigenvectors corresponding to the $k$ largest eigenvalues of $P$.
Let $\Phi^*$ correspond to any set of $k$ eigenvectors of $P$. Then
\begin{align*}
    \ddt{t}\Phi^* = -\left(\Phi^* F^* - P \Phi^*\right) {F^*}^\top = 0.
\end{align*}

To show that all non top-$k$ eigenspaces are asymptotically unstable critical points of the differential equation defined by the gradient flow of $\Phi$.
To show this, we aim to show that there exists an eigenvector of the Jacobian with an eigenvalue larger than $0$.
For this, we construct the directional derivative at the critical point.
The directional derivative is the Jacobian vector product, which allows us to circumvent the need to work with higher order tensor derivatives.
We then proceed to show that there exists a direction which corresponds to the eigenvector of the Jacobian with positive eigenvalue.
This concludes the proof.
This technique closely follows the one used by \citet{lelan2023bootstrapped}.

Assume $\mathrm{span}\{\Phi^*\} \neq \mathrm{span}\{e_1,\dots,e_k\}$. This implies that there exists at least one eigenvector $e_j \in \{e_1,\dots,e_k\}$ and $e_j \notin \mathrm{span}\{\Phi^*\}$, with corresponding eigenvalues $\lambda_j$.

Let $D_\Delta$ be the directional derivative of $\ddt{t} \Phi |_{\Phi=\Phi^*}$ in the direction $\Delta$.
We construct the directional derivative using the product rule (terms colored for ease of reading),
\begin{align*}
D_\Delta \ddt{t} \Phi |_{\Phi=\Phi^*} = & -D_\Delta \left(\left(\Phi F^* - P \Phi\right) {F^*}^\top\right)|_{\Phi=\Phi^*}\\
     = & -{\color{uoftred}D_\Delta \left(\Phi F^* - P \Phi\right)|_{\Phi=\Phi^*}} {F^*}^\top - \left(\Phi^* F^* - P \Phi^*\right) {\color{uoftmagenta}D_\Delta {F^*}|_{\Phi=\Phi^*}}^\top.
\end{align*}

For the directional derivative, we only consider directions that are orthogonal to $\Phi^*$, so ${\Phi^*}^\top\Delta = 0$. Then $\underbrace{P{\Phi^*} = {\Phi^*} A}_{\text{subspace condition}} \implies \Delta^\top P \Phi^* = 0.$
For the derivative with regard to $F^*$, we have
\begin{align*}
    {\color{uoftmagenta}D_\Delta F^*|_{\Phi = \Phi^*}} =& D_\Delta \left(\Phi^\top \Phi\right)^{-1}\Phi^\top P \Phi\nonumber\\
    =& \left(D_\Delta \left(\Phi^\top \Phi\right)^{-1} \right) {\Phi^*}^\top P {\Phi^*} + \left({\Phi^*}^\top {\Phi^*}\right)^{-1} \left(D_\Delta \Phi^\top P \Phi  \right)\nonumber\\
    =& -\underbrace{(\Delta^\top \Phi^* + {\Phi^*}^\top \Delta)}_{=0}\left({\Phi^*}^\top {\Phi^*}\right)^{-2}{\Phi^*}^\top P {\Phi^*} + \left({\Phi^*}^\top {\Phi^*}\right)^{-1}\big(\underbrace{\Delta^\top P {\Phi^*}}_{=0} + {\Phi^*}^\top P \Delta\big)\nonumber\\
    =& \left({\Phi^*}^\top {\Phi^*}\right)^{-1}\left({\Phi^*}^\top P \Delta\right). \label{eq2}
\end{align*}
Therefore, the first term in the second line is dropped, as well as the first term of the final summand.

Note that $F^* = \left({\Phi^*}^\top{\Phi^*}\right)^{-1} {\Phi^*}^\top P^\pi {\Phi^*} = \underbrace{\left({\Phi^*}^\top{\Phi^*}\right)^{-1} {\Phi^*}^\top {\Phi^*}}_{=I}\, \mathrm{diag}(\Lambda_i) = \mathrm{diag}(\Lambda_i)$, where $\Lambda_i$ is the set of eigenvalues corresponding to the eigenvectors in $\Phi^*$ and $\mathrm{diag}(\Lambda_i)$ is the diagonal matrix of eigenvalues corresponding to those eigenvectors spanned by $\Phi^*$.

This allows us to compute the remaining derivative,
\begin{align*}
    &{\color{uoftred}D_\Delta \left(\Phi F^* - P \Phi\right)|_{\Phi=\Phi^*}}\, \mathrm{diag}(\Lambda_i)\\
    = & \left(\Delta\, \mathrm{diag}(\Lambda_i) + \Phi^*{\color{uoftmagenta} D_\Delta F^*}  - P\Delta \right)\,\mathrm{diag}(\Lambda_i)\\
    = & \left(\Delta\, \mathrm{diag}(\Lambda_i) + \Phi^*\left({\Phi^*}^\top {\Phi^*}\right)^{-1}{\Phi^*}^\top P \Delta  - P\Delta \right)\,\mathrm{diag}(\Lambda_i),
\end{align*}
where we use the fact that $D_\Delta P\Phi^* = P D_\Delta \Phi^* = P \Delta$.

Finally, as $\Phi^*F^* = \Phi^* \mathrm{diag}(\Lambda_i) = P^\pi \Phi^*$, we obtain
\begin{align*}
 D_\Delta \ddt{t} \Phi |_{\Phi=\Phi^*} =   -\left(\Delta\, \mathrm{diag}(\Lambda_i) + {\Phi^*}\left({\Phi^*}^\top \Phi^*\right)^{-1}{\Phi^*}^\top P \Delta  - P\Delta \right)\mathrm{diag}(\Lambda_i) - \underbrace{\left(\Phi^* F^* - P \Phi^*\right)}_{=0\,\text{as shown}}\left({\Phi^*}^\top P \Delta\right)^\top.
\end{align*}

By the definition of the directional derivative as the Jacobian-vector product, we can now assert
\begin{align*}
    \left(\ddt{\Phi} \ddt{t}\Phi |_{\Phi=\Phi^*}\right) \Delta = - \left(\Delta \mathrm{diag}(\Lambda_i) + \left(\Phi^*\left({\Phi^*}^\top {\Phi^*}\right)^{-1}{\Phi^*}^\top - I\right) P \Delta\right)\mathrm{diag}(\Lambda_i).
\end{align*}

What remains to be shown is that there exist a direction which corresponds to a positive eigenvalue of the Jacobian of the dynamics.

Choose $\Delta = v_j u^\top$. Let $v_j$ be an eigenvector not in the span of $\Phi^*$ but in the top-k eigenvectors. Let  $\lambda_j$ be the corresponding eigenvalue. By our assumption before, there exist at least one eigenvalue $\lambda_i \in \Lambda_i$ so that $\lambda_j > \lambda_i$.

Note that $\left(\Phi^*\left({\Phi^*}^\top {\Phi^*}\right)^{-1}{\Phi^*}^\top - I\right) P v_j = \Big(\Phi\left({\Phi^*}^\top {\Phi^*}\right)^{-1}\underbrace{{\Phi^*}^\top v_j}_{0\, \text{by construction}} - I v_j\Big) \lambda_j = -\lambda_j v_j$ and therefore $\left(\Phi^*\left({\Phi^*}^\top \Phi^*\right)^{-1}{\Phi^*}^\top - I\right) P \Delta = -\Delta \lambda_j I.$

To simplify notation, we will write $\Lambda_i$ for $\mathrm{diag}(\Lambda_i)$ from now on as there is no risk of confusion.

\begin{align*}
    -\left(\Delta \Lambda_i + \left(\Phi^*\left({\Phi^*}^\top \Phi^*\right)^{-1}{\Phi^*}^\top - I\right) P \Delta\right)\Lambda_i = &- \Delta \left(\Lambda_i - \lambda_j I\right) \Lambda_i\\
    = &-\Delta \left(\Lambda_i^2 - \lambda_j \Lambda_i\right) \\
    = &\Delta \left(\lambda_j \Lambda_i - \Lambda_i^2\right).
\end{align*}

We can now choose $u$ so that it is any eigenvector of $\left(\lambda_j \Lambda_i - \Lambda_i^2\right)$. As this is a diagonal matrix, it is easy to see that if $\lambda_j > \lambda_i$ for any $\lambda_i$, the matrix will have a positive eigenvalue, meaning there exists a direction in which the critical point is unstable.

\end{proof}

\textbf{Case 2: Non-invariant subspace cases:} Not all critical points lie in invariant subspaces.
One such an alternative critical point is the case of $\Phi^\top P \Phi = 0$, and more generally, for each set of column vectors $\phi_i$ in $\Phi$, they needs to either be mapped to an invariant or an orthogonal subspace by $P^\pi$ to be stable.
In the orthogonal case, the Jacobian at the critical point becomes 0, meaning no conclusion about stability can be drawn from this analysis.

We leave further analysis of non invariant subspace critical points open for future work.
We do however conjecture that the non invariant subspace critical points are also saddle-points or unstable solutions of the ODE, following the experimental analysis by \citet{tang2022understanding}.

\textbf{Negative eigenvalues:} In case the matrix has negative eigenvalues, the stability conditions in the final step of the proof change. The matrix $\lambda_j\Lambda_i - \Lambda_i^2$ will not have negative eigenvalues if $\lambda_i$ is negative but $\lambda_j$ is positive. The ranking of stable points follows this slightly un-intuitive ordering: all negative eigenvalues sorted by absolute value followed by all positive eigenvalues sorted by absolute value.

\ReconstructionStationaryPoints*

\begin{proof}
We first show that under the two timescale scenario, $F$ is stationary and therefore does not change the span of the critical points.

Due to the assumption of the two-timescale scenario, we compute $\Psi^*$ by solving the linear regression problem,
\begin{align*}
    &&\ddt{\Psi^*} \lVert \Phi F\Psi^* - P \rVert_F^2 &= F^\top {\Phi}^\top ({\Phi}F\Psi^* - P)\\
    &&0 &=  F^\top {\Phi}^\top ({\Phi}F\Psi^* - P)\\
    \Leftrightarrow && B^* &= \left(F^\top {\Phi}^\top {\Phi}F\right)^{-1} F^\top \Phi^\top P = F^{-1}\left(\Phi^\top \Phi\right)^{-1} \Phi^\top P.
\end{align*}

Plugging this solution back into the original equation, 
\begin{align}
    \lVert \Phi F\Psi^* - P \rVert_F^2 &= \lVert \Phi FF^{-1}\left(\Phi^\top \Phi\right)^{-1} \Phi^\top P - P \rVert_F^2\\
    &= \lVert \Phi \left(\Phi^\top \Phi\right)^{-1} \Phi^\top P - P \rVert_F^2,
\end{align}
we notice that $F$ cancels.
Therefore, the optimality conditions for $A$ follow from the Eckart-Young theorem, as presented in \autoref{lem:rrr}
\end{proof}

\ReparameterizationInvariance*

\begin{proof}
We first write out all losses with the observation matrix $\gO$. The reward function is assumed to not change under the introduction of $\gO$, therefore we do not multiply $\gO$ to $x^\top r$.

\begin{align*}
    L_{\text{rec}}(\Phi,F,\Psi) =& \EEX{x \sim \mathcal{D}}{\ltwonorm{x^\top \gO \Phi F \Psi - x^\top {P^\pi}\gO}^2} = {\ltwonorm{\gO \Phi F \Psi - {P^\pi}\gO}^2}, \\
    L_{\text{lat}}(\Phi,F) =& \EEX{x \sim \mathcal{D}}{\ltwonorm{x^\top \gO \Phi F - \left[x^\top {P^\pi} \gO \Phi\right]_{\mathrm{sg}}}^2} = \ltwonorm{\gO \Phi F - \left[{P^\pi} \gO \Phi\right]_{\mathrm{sg}}}^2 \\
    L_{\text{td}}(\Phi,F) =& \EEX{x \sim \mathcal{D}}{\ltwonorm{x^\top\gO\Phi \hat{V} - \left[x^\top \left(r + \gamma P^\pi \gO\Phi \hat{V}\right)\right]_{\mathrm{sg}}}^2} = {\ltwonorm{\gO\Phi \hat{V} - \left[\left(r + \gamma P^\pi \gO\Phi \hat{V}\right)\right]_{\mathrm{sg}}}^2}.
\end{align*}

Note that in the cases of $L_{\text{lat}}$ and $L_\text{TD}$, all occurrences of $\gO$ are multiplied by $\Phi$.
Therefore the corresponding gradient flows are reparameterizations as defined in \autoref{lem:stability}, and the proof follows directly.
\end{proof}

\ReparameterizationInvarianceObs*

\begin{proof}
    
As before, note that $L_\mathrm{rec}^\gO$ is of the form $\lVert \gO AXB - P\gO \rVert_F^2$, with $A \in \vecR{n}{k}$, $X \in \vecR{k}{k}$, and $B \in \vecR{k}{n}$.

Due to the assumption of the two-timescale scenario, we compute $\Psi^*$ by solving the linear regression problem,
\begin{align*}
    &&\ddt{\psi} \lVert \gO \Phi F \Psi - P \gO \rVert_F^2 &= F^\top \Phi^\top \gO^\top (\gO \Phi F \Psi - P)\\
    &&0 &=  F^\top \Phi^\top \gO^\top (\gO \Phi F \Psi^* - P) = 0 \\
    \Leftrightarrow &&
    \Psi^* &= \left(F^\top \Phi^\top \gO^\top \gO \Phi F\right)^{-1} F^\top \Phi^\top \gO^\top P.
\end{align*}

Substituting into $L_\mathrm{rec}^\gO$, we obtain
\begin{align*}
    \lVert \gO \Phi F \Psi- P \gO \rVert_F^2 & = \lVert \gO \Phi F \left(F^\top \Phi^\top \gO^\top \gO \Phi F\right)^{-1} F^\top \Phi^\top \gO^\top P - P\gO \rVert_F^2 \\
    & = \lVert \gO \Phi \left( \Phi^\top \gO^\top \gO \Phi\right)^{-1} \Phi^\top \gO^\top P - P\gO \rVert_F^2,
\end{align*}
which again implies that $F$ is stationary.

We note that $\lVert \gO \Phi\Psi - P\gO\rVert^2_F$ is the reduced rank regression problem solved in \autoref{lem:rrr} which solution is given by the top-k left and right singular vectors of $\gO^{-1} P \gO$.
\end{proof}

\subsection*{Proofs for Section 5}

\BYOLCombined*
\begin{proof}
        By \autoref{ass:low_rank} there exists a set of $k$ vectors $\phi_1,\dots,\phi_k$ such that $r^\pi \in \text{span}(\phi_1,\dots,\phi_k)$ and $\phi_1,\dots,\phi_k$ span an invariant subspace of $P^\pi$.
        We can choose this set of vectors to orthonormal, e.g. by applying the Gram Schmidt procedure to the eigenvectors $({w_i}_1,\dots{w_i}_m)$. 
        By \autoref{prop:TangResult2} the matrix $\Phi\in \mathbb{R}^{n\times k}$ whose columns are $\phi_1,\dots,\phi_k$ is a critical point for $L_\text{lat}$ and spans an invariant subspace of $P^\pi$.
        In addition, by \autoref{lem:spectrum_rew_value}, $\Phi$ forms a complete basis for the value function $V^\pi$.
        By \autoref{prop:td_critical}, $\Phi$ is also a critical point of $L_\text{TD}$, with $0$ value function approximation error. 
        As $\Phi$ is a critical point for both $L_\text{TD}$ and $L_\text{Lat}$, it is a critical point of $L_\text{TD} + L_\text{Lat}$.

\end{proof}

\ReconsCombined*
\begin{proof}
    Following from \autoref{ass:low_rank} and \cref{lem:spectrum_rew_value}, we have that any critical point $\Phi^*$ of $L_\text{td}$ with perfect value function approximation fulfils $\mathrm{span}(r^\pi)\subseteq \mathrm{span}(\Phi^*)$, as without this condition, $\Phi$ would not have all necessary basis vectors to represent $V^\pi$. Under our low-dimensionality assumption $r^\pi \in \mathrm{span}(w_{i_1},\dots,w_{i_m})$, this condition implies that $\mathrm{span}(w_{i_1},\dots,w_{i_m})\subseteq \mathrm{span}(\Phi^*)$. From \cref{prop:2} we know that any critical point $\Phi^*$ of $L_\text{rec}$ satisfies $\mathrm{span}\,({\Phi^*})=\mathrm{span}\left(\{u_1,\dots,u_k\}\right)$, where $u_1,\dots,u_k$ are the top $k$ left singular vectors of $P^\pi$. Under the assumption that $\mathrm{span}(w_{i_1},\dots,w_{i_m}) \not\subseteq \mathrm{span}(u_1,\dots,u_k)$ these conditions cannot happen simultaneously, and hence no critical point of the joint loss achieves perfect value function reconstruction.
\end{proof}

\topkDistracted*
\begin{proof}
We first note that $\mathrm{span}(V_k) = \mathrm{span}(\{v_i\otimes u_1|\forall i \leq k\})$ follows directly from \autoref{lem:spectrum_kronecker} and \autoref{lem:orth_kronecker}. The eigenvector $u_1 = \vone$ as $M$ is a stochastic matrix.

As $V_k$ is an orthogonal basis, write the projection operation as 
\begin{align*}
V_k V_k^\top \left(r_\gM \otimes \vone\right) &= \frac{1}{m}\left(\vone \otimes \mathrm{orth}(V)\right)\left(\vone \otimes \mathrm{orth}(V)^\top \right)\left(r_\gM \otimes \vone\right) \\
&= \frac{1}{m}\left((\vone \otimes \vone) \otimes (\mathrm{orth}(V)\,\mathrm{orth}(V)^\top)\right) (r_\gM \otimes \vone)\\
&= \frac{1}{m}(\vone \otimes \vone) r_\gM \otimes (\mathrm{orth}(V)\,\mathrm{orth}(V)^\top) \vone \\
&= \sum_{i=1}^m \frac{r_i}{m}\vone.
\end{align*}

\end{proof}

\section{Implementation details for the experiments}
\label{app:empirical}

\begin{table}[]
    \centering
    \begin{tabular}{c|c|c}
        Parameter & MinAtar & DMC \\\hline 
        Initial steps (Random policy) & $5000$ & $5000$  \\
        Env steps per update step & $4$ & $1$\\
        Batch size & $512$ & $512$\\
        Exploration $\epsilon$ & $0.05$ & $0.01$\\
        RL learning rate & $0.0003$ & $0.0003$\\
        Model/decoder learning rate & $0.0003$ & $0.0003$\\
        Encoder learning rate & $0.0001$ & $0.0001$\\
        Target network update interval & $1000$ & n/a\\
        Soft update $\tau$ & n/a & $0.995$ \\
        Discount factor $\gamma$ & $0.99$ & $0.99$\\
        Model forward prediction steps & $4$ & $4$\\
    \end{tabular}
    \caption{Hyper-parameters the RL experiments}
    \label{tab:hyper_minatar}
\end{table}

For the Minatar experiments, we use a simple Double DQN architecture \citep{van2016deep}.
We find that our implementation performs roughly on par with those reported by \citet{young19minatar}, however we changed the network architecture slightly to allow a clear "encoder" and "prediction head" split.

We implement the latent self-prediction loss using a periodically updated copy of the encoder network. This hard update of the encoder target is synchronized with the Q target network update.

Our networks are parameterized as presented in \autoref{tab:net_minatar}

\begin{table}
\begin{center}
\begin{tabular}{c|c|c}
     & ConvLayer & channels=$16$, kernel=$(3,3)$, padding=$0$, stride=$(1,1)$ \\
     Encoder $\Phi$ & ELU activation & -- \\
     & Dense Layer & out\_size=$100$\\
     & ELU activation & -- \\\hline
     & Dense Layer & out\_size=$256$ \\
     Latent Model $F$ & ELU activation & -- \\
     & Dense Layer & out\_size=$100$\\\hline
     & Dense Layer & out\_size=$10\cdot10\cdot16$\\
     Decoder $\Psi$ & ELU activation & -- \\
     & ConvTranspose Layer & kernel=$(3,3)$, padding=$1$,  stride=$(1,1)$\\\hline
     & Dense Layer & out\_size=$256$ \\
     Q head $\hat{V}$ & ELU activation & --\\
     & Dense Layer & out\_size=action\_space
\end{tabular}
\end{center}
\caption{Network architecture for the MinAtar experiments.}
\label{tab:net_minatar}
\end{table}

Relevant hyper-parameters are shown in \autoref{tab:hyper_minatar}.

The random noise matrix is sampled from a Bernoulli distribution with $p(1) = 0.1$.
The distortion matrix was created using a random matrix of size $10\cdot10\cdot\mathrm{channels} \times 10\cdot10\cdot\mathrm{channels}$ with entries independently sampled from a Bernoulli distribution with $p(1) = 0.2$. 
This was chosen to have a similar sparsity in the distraction as in the main observation channels.
Additionally, we verified that the matrix was invertible and $\lVert \gO \rVert_1 < 255$ to ensure that the replay buffer implementation would not overflow.

The DMC results were obtained using the TD3 algorithm \citep{td3} with constant Gaussian noise with standard deviation of $0.01$ added for exploration following \citet{yarats2021image}. In addition to the Q function network, TD3 also requires an actor network. We do not propagate any gradients from the actor network into the encoder, following standard practice in actor-critic learning.

Networks for the DMC implementation are shown in \autoref{tab:net_mujoco}.

\begin{table}
\begin{center}
\begin{tabular}{c|c|c}
     & Dense Layer & out\_size=$256$ \\
     Encoder $\Phi$ & ELU activation & -- \\
     & Dense Layer & out\_size=$256$\\
     & ELU activation & -- \\\hline
     & Dense Layer & out\_size=$256$ \\
     Latent Model $F$ & ELU activation & -- \\
     & Dense Layer & out\_size=$100$\\\hline
     & Dense Layer & out\_size=$10\cdot10\cdot16$\\
     Decoder $\Psi$ & ELU activation & -- \\
     & Dense Layer & out\_size=obs\_dim\\\hline
     & Dense Layer & out\_size=$256$ \\
     Q head $\hat{V}$ & ELU activation & --\\
     & Dense Layer & out\_size=1\\\hline
     & Dense Layer & out\_size=$256$ \\
     Actor head & ELU activation & --\\
     & Dense Layer & out\_size=action\_dim
\end{tabular}
\end{center}
\caption{Network architecture for the DMC experiments.}
\label{tab:net_mujoco}
\end{table}
In the DMC experiments, we used isotropic Gaussian noise for the distracting noise, and a copy of the \emph{humanoid} environment as the distraction.

Full code is available at \url{https://github.com/adaptive-agents-lab/understading_auxilliary_tasks}.

\section{DMC results}
\label{app:mujoco_results}

The DMC environments have a very different observation structure and transition dynamics compared to those of the MinAtar games. The observation spaces are dense and often contain topological discontinuities such as those outlined in \autoref{app:observation_motivation}.

All experiments are repeated across 10 seeds.

\begin{figure}
    \centering
    \includegraphics[width=\textwidth]{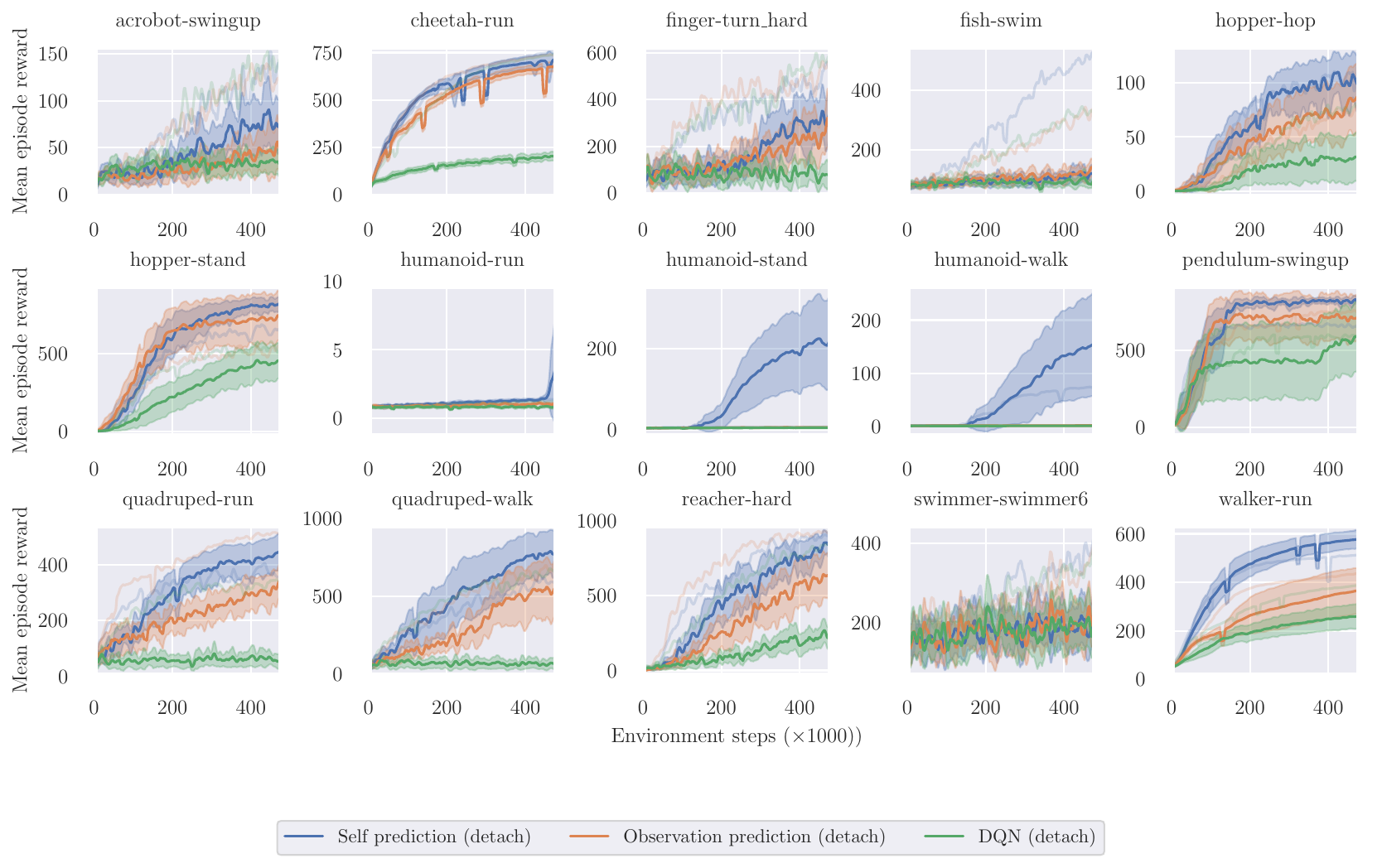}
    \caption{DMC: Stand alone scenario}
    \label{fig:muj_sta}
\end{figure}

\begin{figure}
    \centering
    \includegraphics[width=\textwidth]{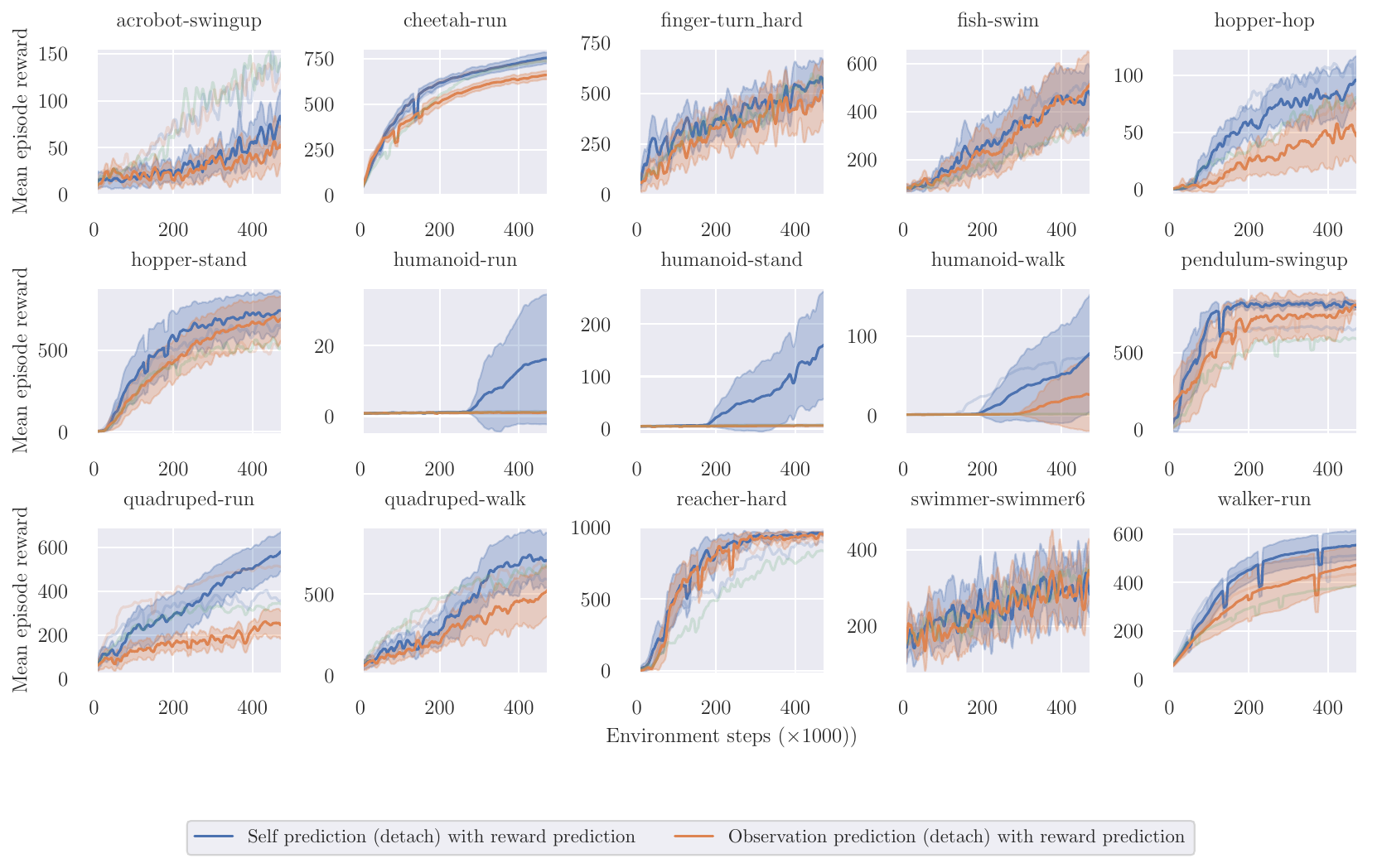}
    \caption{DMC: Auxiliary loss + reward prediction, not TD}
    \label{fig:muj_sta_rew}
\end{figure}

We find slightly different results in these environments compared to MinAtar, especially about the efficacy of the latent self-prediction and observation reconstructions in the stand-alone setting (\autoref{fig:muj_sta} and \autoref{fig:muj_sta_rew}). Overall, latent self-prediction performs much more strongly in these environments compared to the MinAtar experiments, especially when using it as a stand-alone loss.
Curiously, in the only environment where observation prediction shines (quadruped-walk), stand-alone latent self-prediciton and reward prediction seems to outperform all other test settings.
This highlights the second important difference between the MinAtar games and the DMC suite: dense rewards.
We conjecture that most of the differences between observation prediction and latent self-prediction in the DMC suite comes from dense rewards and different topological continuity, but a precise investigation is needed in future work.

Finally, our assumptions about the impact of noise distortions on the efficacy of different loss functions seems to be much more clearly apparent in DMC than in MinAtar. This suggests that the differing primary observation spaces also change how the learning process interacts with the noise, i.e. because the spectral properties of the underlying environments might differ.
Especially in the challenging distraction setting of structured noise, non trivial policies (as measured by a strong performance improvement over a random baseline) can only be observed in 7 out of the 15 environemnts. Of these 7, latent self prediction seems to outperform the other baselines in 3 cases, only falling behind in 1. However, we observed that the pendulum environment has a strong bimodal distribution with some runs completely failing to perform for each experimental scenario, so the number of seeds might be insufficient to disambiguate performance here, as evidenced by the large confidence interval at the $95\%$ level.

Overall, 10 seeds, even though widely used in the community, might not be sufficient to fully represent the performance in the DMC suite, suggesting that more work is needed on the inherent variations of these environments.
We do not present aggregate performance over the environments, as the different reward scales makes a comparison prone to be driven by outliers such as the humanoid environments, in which several algorithms fail to learn at all at 10 seeds.

\begin{figure}
    \centering
    \includegraphics[width=\textwidth]{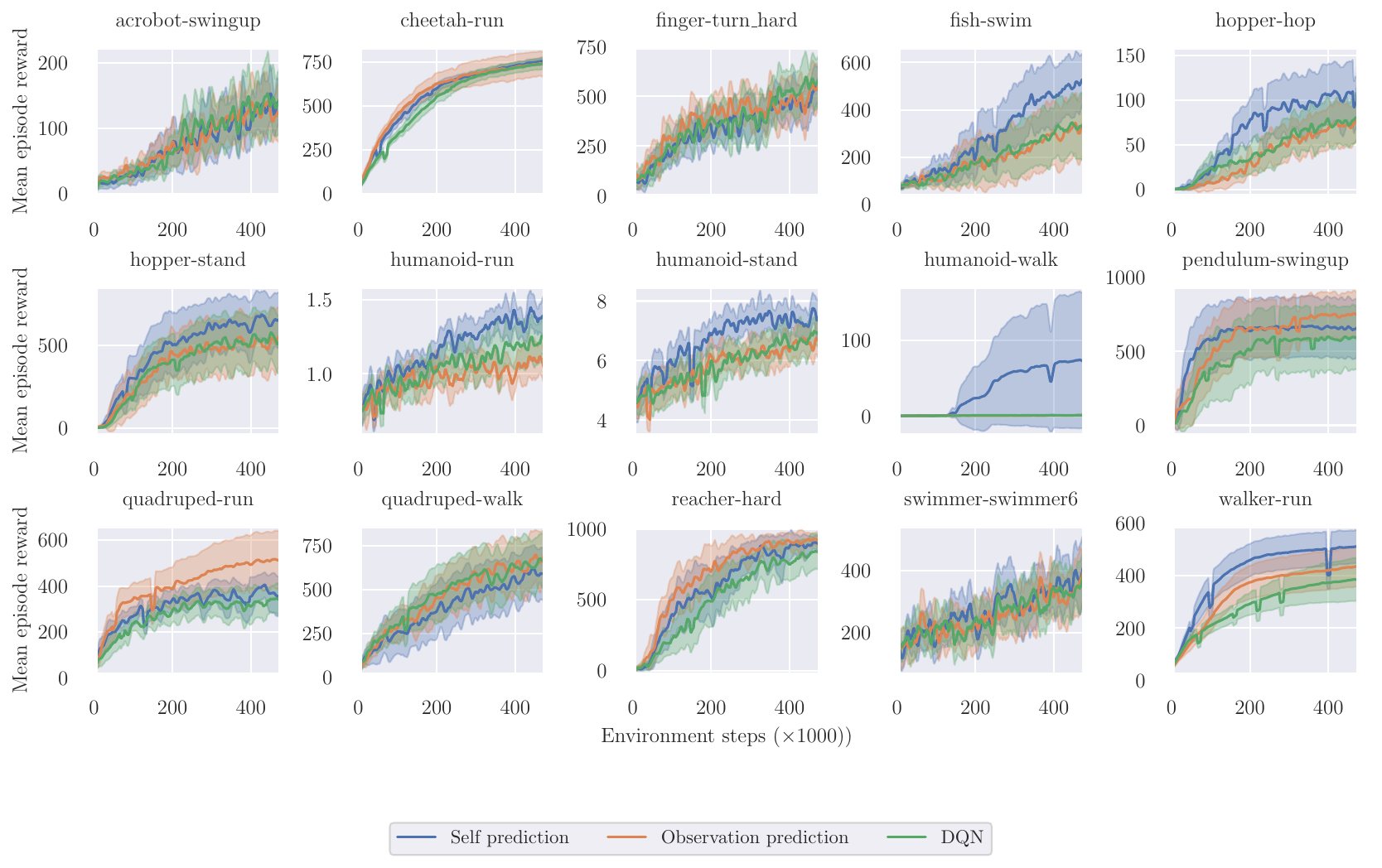}
    \caption{DMC: Auxiliary task scenario}
    \label{fig:muj_aux}
\end{figure}

\begin{figure}
    \centering
    \includegraphics[width=\textwidth]{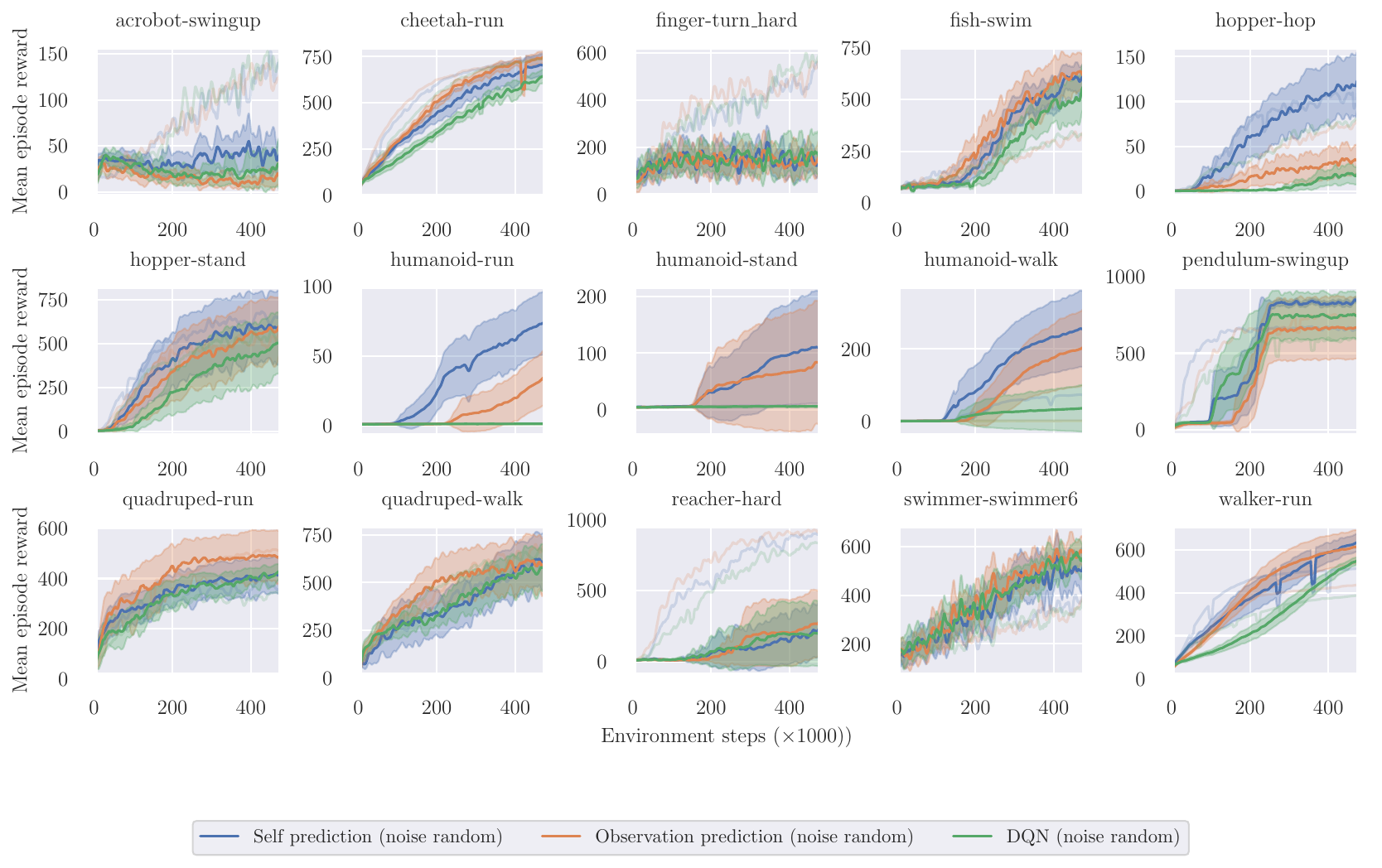}
    \caption{DMC: Random noise distraction}
    \label{fig:muj_ran_noi}
\end{figure}

\begin{figure}
    \centering
    \includegraphics[width=\textwidth]{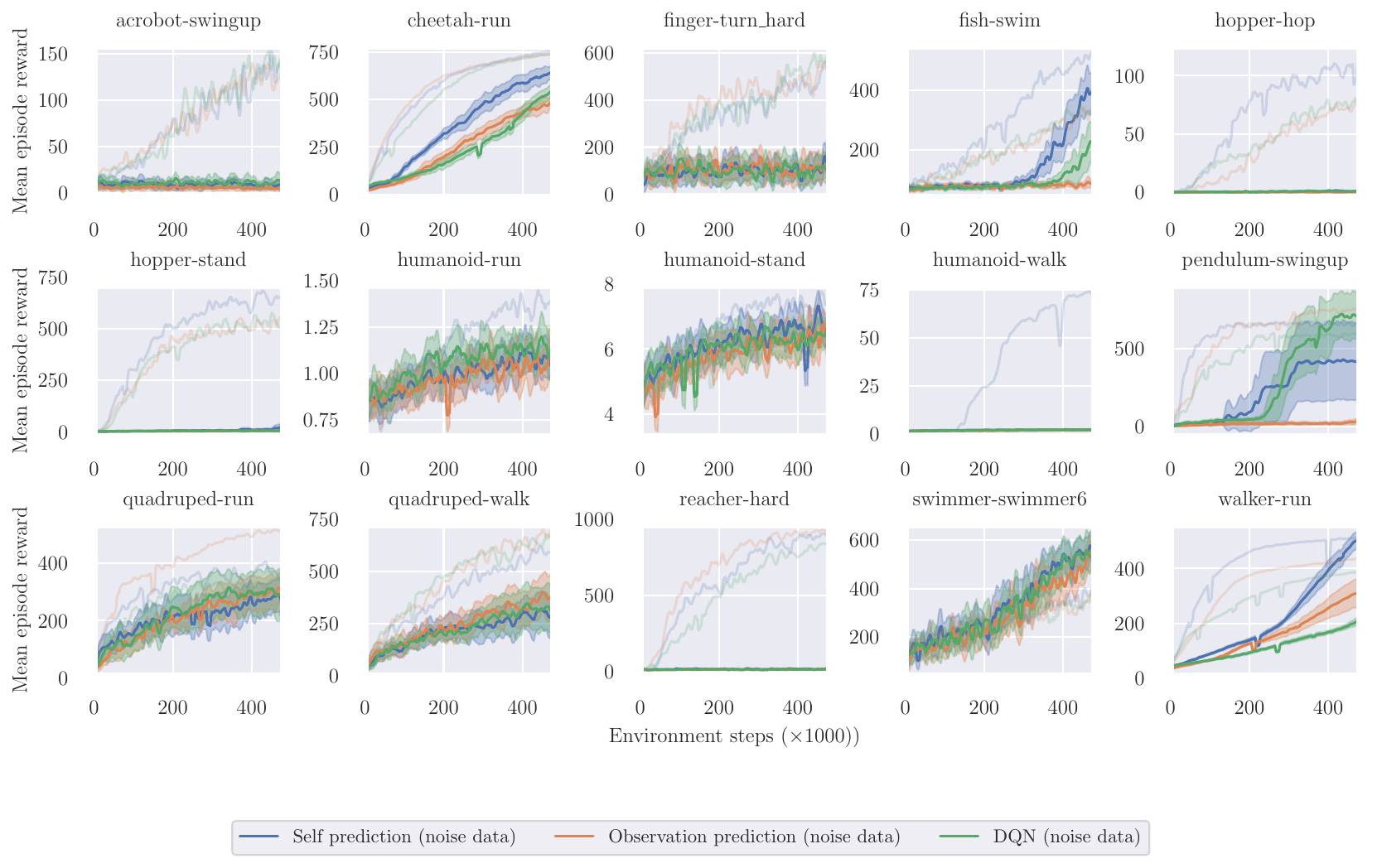}
    \caption{DMC: Structured distraction}
    \label{fig:muj_ran_str}
\end{figure}

\begin{figure}
    \centering
    \includegraphics[width=\textwidth]{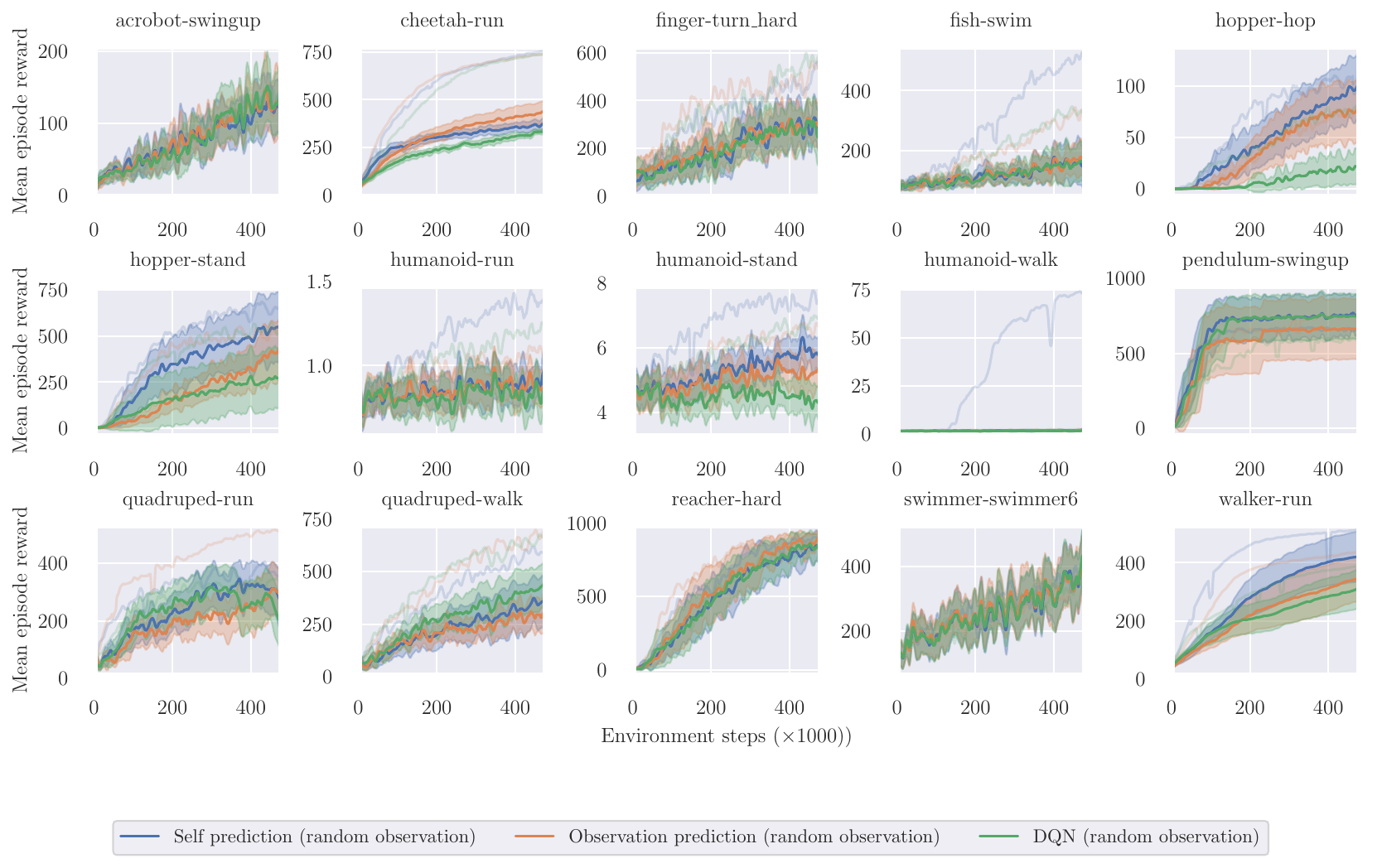}
    \caption{DMC: Observation space distortion}
    \label{fig:muj_dis}
\end{figure}

\section{Limitations}
\label{app:limitations}

While our paper aims to minimize the theory-practice gap with careful experimentation, we nonetheless need to make several assumptions that are both limitations and also potential for additional analysis in future work.
As our aim is to make theory useful and accessible for practitioners, we aim to be very open and clear about our limitations here.

\textbf{Limitations of the analytical framework:} From previous work \citep{tang2022understanding,lelan2023bootstrapped} we inherit the limitation of studying deterministic models in potentially stochastic environments.
This is a necessary limitation, as considering the stochastic equivalents of e.g. the observation prediction loss would render the model and their gradients non-linear due to the introduction of a softmax or similar constraint.
As other works carry the same limitation, we believe that this does not render our work inapplicable, but it does suggest the need for more powerful mathematical tools in future work.

We also conduct all of our theoretical work in the on-policy policy evaluation regime, while our empirical study includes both off-policy policy estimation and policy improvement.
Again, this is a limitation inherited from all related work.
As our theoretical predictions are still validated, we consider this an acceptable limitation, but studying the impact of off-policy samples and shifting policies is an important step for future work.

\textbf{Limitations of the formalism:} Our notion of observation distortion requires unnaturally large observation spaces. Again, this stems from our adherence to a linear framework.
While the observation space of e.g. the MinAtar games is relatively large, depending on the game a 300-1000 dimensional vector, this is still substantially smaller than the total number of states.
Studying these kinds of nonlinear image transformations in more detail would require deviating farther from the previous literature.

In this paper, we aim to (re-)introduce the notion of distraction into the learning dynamics literature, and so we use a relatively simplistic notion of distraction.
Going beyond the independence assumption in the distraction model (due to the Kronecker formulation) and analysing more complex forms of distractions e.g. processes in which the reward-relevant process causally influences the distracting process but not vice versa is an exciting direction for future work.

\end{document}